\documentclass[11pt,letterpaper]{article}
\usepackage{multirow}
\usepackage[utf8]{inputenc}
\usepackage{etex}
\usepackage{xspace,enumerate}
\usepackage{subcaption}
\usepackage[dvipsnames]{xcolor}
\usepackage[T1]{fontenc}
\usepackage[full]{textcomp}
\usepackage[american]{babel}
\usepackage{mathtools}
\usepackage{amsthm}
\newtheorem{theorem}{Theorem}[section]
\newtheorem*{theorem*}{Theorem}

\newtheorem*{proposition*}{Proposition}
\newtheorem{lemma}[theorem]{Lemma}
\newtheorem*{lemma*}{Lemma}
\newtheorem{corollary}[theorem]{Corollary}
\newtheorem*{conjecture*}{Conjecture}

\newtheorem*{fact*}{Fact}

\newtheorem*{hypothesis*}{Hypothesis}

\theoremstyle{definition}
\newtheorem{definition}[theorem]{Definition}
\newtheorem*{definition*}{Definition}

\theoremstyle{remark}

\newtheorem*{claim*}{Claim}

\newtheorem*{remark*}{Remark}

\newtheorem*{observation*}{Observation}

\usepackage[
letterpaper,
top=1.2in,
bottom=1.2in,
left=1in,
right=1in]{geometry}
\usepackage{newpxtext} % T1, lining figures in math, osf in text
\usepackage{textcomp} % required for special glyphs
\usepackage[varg,bigdelims]{newpxmath}
\usepackage[scr=rsfso]{mathalfa}% \mathscr is fancier than \mathcal
\usepackage{bm} % load after all math to give access to bold math
\let\mathbb\varmathbb
\usepackage{microtype}
\usepackage[
 colorlinks=true,
 urlcolor=blue,
 linkcolor=red,
 citecolor=OliveGreen,
 ]{hyperref}
\usepackage[capitalise,nameinlink]{cleveref}
\crefname{lemma}{Lemma}{Lemmas}
\crefname{fact}{Fact}{Facts}
\crefname{theorem}{Theorem}{Theorems}
\crefname{corollary}{Corollary}{Corollaries}
\crefname{claim}{Claim}{Claims}
\crefname{example}{Example}{Examples}
\crefname{algorithm}{Algorithm}{Algorithms}
\crefname{problem}{Problem}{Problems}
\crefname{definition}{Definition}{Definitions}
\usepackage{paralist}
\usepackage{turnstile}
\usepackage{tikz}
\usepackage{caption}
\DeclareCaptionType{Algorithm}
\usepackage{newfloat}
\usepackage{pst-node}

\usepackage{subcaption}
\usepackage{pgfplots}
\pgfplotsset{compat=1.16}
\usepackage{url}

\usepackage{comment}
\def\AA{\mathbf{A}}
\def\BB{\mathbf{B}}

\def\MM{\mathbf{M}}

\def\WW{\mathbf{W}}
\def\CC{\mathbf{C}}

\def\II{\mathbb{I}}
\def\YY{\mathbf{Y}}
\def\XX{\mathbf{X}}
\def\PP{\mathbf{P}}

\def\UU{\mathbf{U}}

\def\VV{\mathbf{V}}

\def\SS{\mathbf{S}}

\def\WW{\mathbf{W}}

\def\MM{\mathbf{M}}

\def\ZZ{\mathbf{Z}}
\def\RR{\mathbf{R}}
\def\II{\mathbf{I}}

\def\S{\mathcal{S}}

\def\R{\mathcal{R}}

\def\cK{\mathcal{K}}

\def\bp{{\bf q}}
\def\bq{{\bf p}}
\def\bu{{\bf u}}
\def\bv{{\bf v}}
\def\bx{{\bf x}}
\def\bw{{\bf w}}
\def\by{{\bf y}}
\def\bz{{\bf z}}
\def\bM{{\bf M}}
\def\bA{{\bf A}}
\def\bP{{\bf P}}
\def\bW{{\bf W}}
\def\bB{{\bf B}}

\def\Null{\mbox{Null}}
\def\Span{\mbox{Span}}
\def\Proj{\mbox{Proj}}

\def\Proj{\mbox{Proj}}
\def\nnz{\textsf{nnz}}

\def\Sig{\mathbf{\Sigma}}

\usepackage{ amssymb }

\def\Span#1{\textbf{Span}\left(#1  \right)}

\def\poly{\textrm{poly}}

\DeclareMathOperator*{\argmax}{argmax}
\DeclareMathOperator*{\argmin}{argmin}

\usepackage[framemethod=TikZ]{mdframed}
\newcounter{Frame}
\newenvironment{Frame}[1][htb]{%
\refstepcounter{Frame}
    \begin{mdframed}[%
        frametitle={#1},
        skipabove=\baselineskip plus 2pt minus 1pt,
        skipbelow=\baselineskip plus 2pt minus 1pt,
        linewidth=1.0pt,
        frametitlerule=true,
        nobreak=true
    ]%
}{%
    \end{mdframed}
}

\title{Learning a Latent Simplex in
Input-Sparsity Time}

\author{Ainesh Bakshi\\
Carnegie Mellon University\\
abakshi@cs.cmu.edu
\and
Chiranjib Bhattacharyya \\
Indian Institute of Science\\
chiru@iisc.ac.in
\and
Ravi Kannan \\
Microsoft Research India\\
kannan@microsoft.com
\and
David P. Woodruff\\
Carnegie Mellon University\\
dwoodruf@cs.cmu.edu
\and
Samson Zhou\\
Carnegie Mellon University\\
samsonzhou@gmail.com
}

\begin{document}
\maketitle

%!TEX root = main.tex

\begin{abstract}
We consider the problem of learning a latent $k$-vertex simplex $\cK\subset\mathbb{R}^d$, given access to  $\AA\in\mathbb{R}^{d\times n}$, which can be viewed as a data matrix with $n$ points that are obtained by randomly perturbing latent points in the simplex $\cK$ (potentially beyond $\cK$). A large class of latent variable models, such as adversarial clustering, mixed membership stochastic block models, and topic models can be cast as learning a latent simplex. Bhattacharyya and Kannan (SODA, 2020) give an algorithm for learning such a latent simplex in time roughly $O(k\cdot\nnz(\AA))$, where $\nnz(\AA)$ is the number of non-zeros in $\AA$. We show that the dependence on $k$ in the running time is unnecessary given a natural assumption about the mass of the top $k$ singular values of $\AA$, which holds in many of these applications. Further, we show this assumption is necessary, as otherwise an algorithm for learning a latent simplex would imply an algorithmic breakthrough for spectral low rank approximation. 

At a high level, Bhattacharyya and Kannan provide an adaptive algorithm that makes $k$ matrix-vector product queries to $\AA$ and each query is a function of all queries preceding it. Since each matrix-vector product requires $\nnz(\AA)$ time, their overall running time appears unavoidable.
Instead, we obtain a low-rank approximation to $\AA$ in input-sparsity time and show that the column space thus obtained has small $\sin\Theta$ (angular) distance to the right top-$k$ singular space of $\AA$. Our algorithm then selects $k$ points in the low-rank  subspace with the largest inner product (in absolute value) with $k$ carefully chosen random vectors. By working in the low-rank subspace, we avoid reading the entire matrix in each iteration and thus circumvent the $\Theta(k\cdot\nnz(\AA))$ running time. 
\end{abstract}

\section{Introduction}

We study the problem of learning $k$ vertices $\MM_{*,1},\ldots,\MM_{*,k}$ of a latent $k$-dimensional simplex $\cK$ in $\mathbb{R}^d$ using $n$ data points generated from $\cK$ and then possibly perturbed by a stochastic, deterministic, or adversarial source before given to the algorithm. 
In particular, the resulting points observed as input data could be heavily perturbed so that the initial points may no longer be discernible or they could be outside the simplex $\cK$. 
Recent work of Bhattacharyya and Kannan \cite{BK20} unifies several stochastic models for unsupervised learning problems, including $k$-means clustering~\cite{celeux1992classification, ghahramani1996algorithm,weber2003clustering, witten2010framework, duan2020latent}, topic models~\cite{blei2003modeling, steyvers2007probabilistic,blei2006correlated, Blei12, arora2013practical}, mixed membership stochastic block models~\cite{A08,miller2009nonparametric,xing2010state,fu2009dynamic, A14, li2016scalable, fan2016copula} and Non-negative Matrix Factorization~\cite{arora2013icml,gillis2014fast,gillis2020nonnegative}  under the problem of learning a latent simplex. 
In general, identifying the latent simplex can be computationally intractable. 
However many special applications do not require the full generality. 
For example, in a mixture model like Gaussian mixtures, the data is assumed to be generated from a convex combination of density functions. 
Thus, it may be possible to efficiently approximately learn the latent simplex given certain distributional properties in these models. 

Indeed, Bhattacharyya and Kannan  showed that given certain reasonable geometric assumptions that are typically satisfied for real-world instances of Latent Dirichlet Allocation, Stochastic Block Models and Clustering, there exists an $\widetilde{O}(k \cdot \nnz(\AA))$
\footnote{Throughout the paper we use the notation $\widetilde{O}$ to suppress poly-logarithmic factors.} 
time algorithm for recovering the vertices of the underlying simplex. 
We show that, given an additional natural assumption, we can remove the dependency on $k$ and obtain a true input sparsity time algorithm. 
We begin by defining the model along with our new assumption:
\begin{definition}[Latent Simplex Model]
\label{def:latent_simplex}
Let $\MM$ be a $d \times k$ matrix such that $\MM_{*, 1},\MM_{*,2},\ldots ,\MM_{*,k} \in \mathbb{R}^d$ denote the vertices of a $k$-simplex, $\cK$. Let $\PP$ be a $d \times n$ matrix such that $\PP_{*,1}, \PP_{*,2}\ldots \PP_{*,n} \in \mathbb{R}^d$ are $n$ points in the convex hull of $\cK$. Given $\sigma>0$, we observe a $d \times n$ matrix $\AA$, such that $\|\AA - \PP \|_2 \leq \sigma \sqrt{n}$. 
Further, we make the following assumptions on the data generation process:
\begin{enumerate}
    \item \textbf{Well-Separateness. }\label{a1} For all $\ell \in [k]$, $\MM_{*,\ell}$ has non-trivial mass in the orthogonal complement of the span of the remaining vectors, i.e., for all $\ell \in [k]$, 
    $|\textrm{Proj}(\MM_{*,\ell}, \textrm{Null}(\MM \setminus \MM_{*,\ell}) ) | \geq \alpha \max_{\ell} \|\MM_{*,\ell} \|_2$
    where $\textrm{Proj}(x,U)$ denotes the orthogonal projection of $x$ to the subspace $U$ and $\MM \setminus \MM_{*,\ell}$ is the matrix $\MM$ with the $\ell$-th column removed. 
    \item \textbf{Proximate Latent Points. }\label{a2} Given $\delta \in (0,1)$, for all $\ell \in [k]$, there exists a set $\S_{\ell} \subseteq [n]$ such that $|\S_{\ell}| \geq \delta n$ and for all $j\in \S_{\ell}$, $\|\MM_{*,\ell} - \PP_{*,j} \|_2 \leq 4 \sigma/\delta$. 
    \item \textbf{Spectrally Bounded Perturbation.}\label{a3} The spectrum of $\AA - \PP$ is bounded, i.e., for a sufficiently large constant $c$, $\sigma/\sqrt{\delta} \leq \alpha^2 \min_{\ell} \|\MM_{*,\ell} \|_2/ck^{9} $. 
    \item \textbf{Significant Singular Values.}\label{a4} Let $\AA = \sum_{i \in [d]} \sigma_i u_i v_i^{T}$ be the singular value decomposition and let $0< \phi \leq \nnz(\AA)/(n \cdot\poly(k))$. We assume that for all $i \in [k]$, $\sigma_i>\phi\cdot\sigma_{k+1}$ and $\|\AA - \AA_k\|^2_F \leq \phi \|\AA - \AA_k\|^2_2$.  
\end{enumerate}
\end{definition}
These assumptions are natural across many interesting applications; see Section \ref{sec:applications} for more details. 
\cite{BK20} introduced the Well-Separateness \eqref{a1}, Proximate Latent Points \eqref{a2} and Spectrally Bounded Perturbation \eqref{a3} assumptions. 
%They obtain a $\widetilde{O}\left(k\cdot\nnz(\AA) + (n+d)\poly(k)\right)$ time algorithm to recover the vertices of the Latent Simplex. 
We include an additional Significant Singular Values assumption \eqref{a4}, which is crucial for obtaining a faster running time; we discuss this in more detail below.
%Given an instance from the aforementioned model, the goal is to recover vertices of the latent simplex $\cK$.  
Our main algorithmic result can then be stated as follows:

\begin{theorem}[Learning a Latent Simplex in Input-Sparsity Time]
\label{thm:input_sparsity_simplex}
Given $k\ge 2 $ and $\AA \in \mathbb{R}^{d\times n}$ from the Latent Simplex Model (Definition \ref{def:latent_simplex}), there exists an algorithm that runs in $\widetilde{O}\left(\textsf{nnz}(\AA) + (n+d)\poly(k/\phi) \right)$ time to output subsets $\AA_{\R_{1}},\ldots,\AA_{\R_{k}}$ such that upon permuting the columns of $\MM$, with probability at least $1-1/\Omega(\sqrt{k})$, for all $\ell\in[k]$,  we have $\|\AA_{\R_{\ell}}-\MM_{*,\ell}\|_2\le 300k^4 \sigma/(\alpha\sqrt{\delta})$.
\end{theorem}

Our result implies faster algorithms for various stochastic models that can be formulated as special cases of the Latent Simplex Model, including Latent Dirichlet Allocation for Topic Modeling, Mixed Membership Stochastic Block Models and Adversarial Clustering. 
We summarize the connections to these applications below. 
We describe our algorithm and provide an outline to our analysis; we defer all formal proofs to the supplementary material. 

\section{Connection to Stochastic Models}
\label{sec:applications}
We first formalize the connection between the Latent Simplex Model (Definition \ref{def:latent_simplex}) and numerous stochastic models. 
In particular, we show that topic models like Latent Dirichlet Allocation (LDA), Stochastic Block Models and Adversarial Clustering can be viewed as special cases of the Latent Simplex Model. We also show how our assumptions are natural in each of these applications. 

\subsection{Topic Models}
Probabilistic Topic Models attempt to identify abstract topics in a collection of documents by discovering latent semantic structure \cite{blei2003modeling,blei2006dynamic, hoffman2010online,zhu2012medlda,Blei12}.  Each document in the corpus is represented by a bag-of-words vectorization with the corresponding word frequencies. The standard statistical assumption is that the generative process for the corpus is a joint probability distribution over both the observed and hidden random variables. The hidden random variables can be interpreted as representative documents for each topic. The goal is to then design algorithms that can learn the underlying topics. 
The topics can be viewed geometrically as  $k$ latent vectors $\MM_{*,1}, \MM_{*,2},\ldots ,
\MM_{*,k}\in \mathbb{R}^d$, where $d$ is the size of the dictionary and $\MM_{i,\ell}$ is the expected frequency of
word $i$ in topic $\ell$. Since each vector $\MM_{*,\ell}$ represents a probability distribution, $\sum_i \MM_{i,\ell} =1$. Let $\MM$ be the corresponding $d \times k$ matrix. 
One important stochastic model is Latent Dirichlet Allocation (LDA) \cite{lda}, where each document consists of $m$ words is generated as follows :
\begin{itemize} 
\item For all $\ell \in [k]$, we pick topic weights $\WW_{j, \ell} \sim \textsf{Dir}(1/k)$,  where $\textsf{Dir}(1/k)$ is the Dirichlet distribution over the unit simplex.  
The topic distribution of document $j$ is decided by the topic weights, $\WW_{j, \ell}$, and given by $\PP_{*,j}=\sum_{\ell \in [k]} \WW_{j, \ell} \cdot \MM_{*, \ell}$, where
$\PP_{*,j}$ are latent points.
\item We then generate the $j$-th document with $m$ words by taking i.i.d. samples from $\textsf{Mult}(\PP_{*,j})$, the multinomial distribution with $\PP_{*,j}$ as the probability vector. The resulting document observed is denoted by the vector $\AA_{*,j}$, where for all $i\in[d]$
$\AA_{i,j}=\frac{1}{m}\sum_{t=1}^{m} \XX^{(t)}_{ij},$, such that $\XX^{(t)}_{ij} \sim \textsf{Bern}(\PP_{ij})$,
where $\XX^{(t)}_{ij} = 1$ if the $i$-th word was chosen in the $t$-th draw while generating the $j$-th document, and $0$ otherwise.
\end{itemize}
The data generation process of LDA can be viewed as a special case of the Latent Simplex Model, where the $j$-th document is the data point $\AA_{*,j}$ generated from the stochastic vector $\PP_{*,j}$, a point in the simplex $\cK$. The vertices of the simplex are the $k$ topic vectors $\MM_{*,1},\ldots,\MM_{*,k}$; the goal is then to recover the vertices of $\cK$. 
\cite{BK20} remark that the Well-Separateness condition holds for LDA if we assume a Dirichlet prior on $\MM$.
We note that while $\cK$ is a $k$-dimensional simplex, $d \ll k$ and the observed points need not lie inside the simplex. On the contrary, \cite{BK20} show that the data often lies significantly outside of $\cK$. However, they show that the smoothed simplex obtained by taking the averages of all $\delta n$ sized subsets of observed points results in a polytope $K_{\S}$ that is close to $\cK$. 

We formally justify our assumptions below.
\begin{lemma}[LDA as a Latent Simplex]
\label{lem:lda_as_simplex}
Given $\AA, \PP, \MM$ following the LDA model as described above, such that for all $\ell \in [k]$, $\|\MM_{*,\ell}\|_2 = \Omega(1)$, $m, n = \Omega(\poly(k/\alpha))$ and $\delta = c\sigma/\sqrt{k}$, assumptions \eqref{a2},\eqref{a3} and \eqref{a4} from 
%Proximate Latent Points, Spectrally Bounded Perturbation, and Significant Singular Values assumptions from
Definition \ref{def:latent_simplex} are satisfied with high probability. 
\end{lemma}

\begin{proof}
Assumptions \eqref{a2} and \eqref{a3} follow from Lemma 7.1  in \cite{BK20}. 
By Claim 8.1 in \cite{BK20}, $\sigma_k(\AA) \geq c\alpha  \sqrt{\delta/k} \min_{\ell}\MM_{*,\ell}$. 
Each column of $\AA$ sums to $1$, so $\|\AA \|^2_F = O(n)$ and $\sigma_k(\AA) \geq \alpha \sqrt{ \delta /k}\|\AA \|_F $. 
Since $\|\AA-\PP\|_2\le\sigma\sqrt{n}$ by definition of $\sigma$, and $\PP$ consists of $n$ point in the convex hull of $k$ points and thus $\sigma_{k+1}(\PP)=0$, we have $\sigma_{k+1}(\AA) \leq \sigma_{k+1}(\PP) + \|\AA - \PP \|_2 \leq \sigma \sqrt{n} \leq \sigma \|\AA\|_F$. 
Thus if $\sigma \leq \alpha \sqrt{\delta}/\poly(k)$ for a large enough $\poly(k)$, our Significant Singular Values assumption holds.
\end{proof}

% We give the following further justification for assumption~\eqref{a4} in the supplementary material, showing that removing the assumption would result in a faster algorithm for spectral low rank approximation, partially resolving the first open question of \cite{woodruff2014sketching}.
% \begin{theorem}[Spectral LRA and Learning a Simplex (informal)]
% There exists a distribution  such that learning a latent simplex in input sparsity time implies a constant factor spectral low-rank approximation algorithm in input sparsity time for the instance.
% \end{theorem}

\subsection{Mixed Membership Stochastic Block Models}
\label{section:MMBM}

The Stochastic Block Model is a well-studied stochastic model for generating random graphs, where the vertices are partitioned into $k$ communities and edges within each community are more likely to occur than edges across communities. 
Given communities $C_1, C_2,\ldots C_k$, there exists a $k\times k$ symmetric latent matrix $\bB$, where, $\BB_{\ell_1,\ell_2}$ is the probability that there exists an edge between vertices in  $C_{\ell_1}$ and $C_{\ell_2}$. 
The MMBM can be formalized as the following stochastic process:

\begin{itemize} 
    \item For $j \in [n]$, vertex $j$ picks a probability vector $\WW_{*, j} \in \mathbb{R}^k$ representing community membership probabilities that sum to $1$, i.e., $\WW_{i,j}\sim \textsf{Dir}(1/k)$ for all $i\in [k]$. 
    \item For all pairs $(j_1, j_2) \in [n]$, vertex $j_1$ picks a community $\ell_1$ proportional to $\textsf{Mult}(\WW_{*, j_1})$ and $j_2$ picks a community $\ell_2$ proportional to $\textsf{Mult}(\WW_{*, j_2})$. The edge $(j_1, j_2)$ is included in the graph with probability $\BB_{\ell_1, \ell_2}$. Since $\sum_{\ell_1, \ell_2} \WW_{\ell_1, j_1} \BB_{\ell_1, \ell_2} \WW_{\ell_2, j_2}$ represents the edge probability of the edge $(j_1, j_2)$, the latent variable matrix $\PP$ of edge probabilities can be represented as $\PP = \WW^T \BB \WW^T$. 
\end{itemize}

However, our reduction is not straightforward since now $\bP$ depends quadratically on $\bW$ and the only polynomial time algorithms for $\bB$ directly rely on semidefinite programming. Further, they require non-degeneracy assumptions in order to compute a tensor decomposition provably in polynomial time \cite{anima14,Hop17}.
However, we can pose the problem of recovery of the $k$ underlying communities differently and first pick at random a subset $V_1 \subset [n]$ of $d$ vertices and represent the $\ell$-th community by a $d$-dimensional vector that represents the probabilities of
vertices in $[n] \setminus V_1$ belonging to community $\ell$ and having an edge with each of the $d$ vertices in $V_1$. 
We now define $\WW_{(1)}$ to be a $k \times d$ matrix representing the fractional membership of weights of vertices in $V_1$ and $\WW_{(2)}$ to be the analogous $k \times n$ matrix for vertices in $[n]\setminus V_1$. Observe that the probability matrix $\PP$ can now be represented as $\WW^T_{(1)} \BB \WW_{(2)}$. 

The reduction to the Latent Simplex Model can now be stated as follows: given a data matrix $\AA$ which is the adjacency matrix of the community graph, and the latent variable matrix $\PP$, recover the simplex $\bM = \WW_{(1)}^T \BB$. Further, 
\cite{A08} assumes that each column of $\bW_{(2)}$ is picked from the Dirichlet distribution with parameter $1/k$. Combined with tools from random matrix theory \cite{vers}, \cite{BK20} (Lemma 7.2) shows that the Proximate Latent Points and Spectrally Bounded assumptions hold for Stochastic Block Models.
%(Lemma 7.2 in \cite{BK20}). 
As for the Significant Singular Values assumption, it is satisfied when $\sigma$ is a small enough polynomial in $k$. 

\paragraph{Justifying Significant Singular Values.} We give the following further justification for assumption~\eqref{a4} in Section \ref{app:mixed:reduction}: a faster algorithm only using the assumptions appearing in \cite{BK20} would imply an algorithmic breakthrough for spectral low-rank approximation and partially resolve the first open question of \cite{woodruff2014sketching}.
%removing the assumption would result in a faster algorithm for spectral low rank approximation, partially resolving the first open question of \cite{woodruff2014sketching}.
\begin{theorem}[Spectral LRA and Learning a Simplex (informal)]
There exists a distribution over instances such that learning a latent simplex in $o(nnz(\AA)\cdot k)$ time with good probability implies a constant factor spectral low-rank approximation algorithm in the same running time.
\end{theorem}

\subsection{Adversarial Clustering}\label{sec:advinf}
We consider clustering problems that arise naturally from stochastic mixture models such as Gaussian, Mallows, categorical and so on \cite{sanjeev2001learning,vempala2004spectral,lu2011learning,charikar2017learning,diakonikolas2018list,liu2018efficiently}. 
We can then formulate such a clustering problem in the Latent Simplex Model as follows:  
Given $n$ data points $\AA_{*,1},\AA_{*,2},\ldots ,\AA_{*,n}\in  \mathbb{R}^d$, such that the data is a mixture of $k$ distinct clusters,  $
\CC_1,\CC_2,\ldots ,\CC_k$, with means $\MM_{*,1},\MM_{*,2},\ldots ,\MM_{*,k}$, the goal is to approximately learn the means. Further, we can set the $n$ latent vectors $\PP_{*,j}$ to denote the mean of the cluster point $\AA_{*,j}$ belongs to, and thus $\PP_{*,j}\in\{ \MM_{*,1},\MM_{*,2},\ldots,
\MM_{*,k}\}$. Prior work of 
\cite{KK10} and \cite{AS12} shows that if the minimum cluster size if $\delta n$ and for all $ \ell \neq \ell'$,  $\|\MM_{*,\ell}-\MM_{*,\ell'}\|\geq ck\frac{\sigma}{\sqrt\delta}$
the $\MM_{*, \ell}$ can be
found within error $O(\sqrt k\sigma/\sqrt\delta)$.

However, the aforementioned algorithms are not robust to adversarial perturbations. Therefore, we describe the perturbations we can handle in the Latent Simplex Model. The adversarial model is the same as the one considered in \cite{BK20}. The adversary is allowed to selected a subset $S_{\ell}$ of each cluster $\CC_{\ell}$ of cardinality at most $\delta n$ and perturb each point $\AA_{*,j}$ for $j \in S_{\ell}$ by $\Delta_j$ such that :
\begin{itemize} 
\item $\PP_{*,j}+\Delta_j $ is still in the Convex Hull of $\text{ } \MM_{*,1},\MM_{*,2},\ldots ,\MM_{*,k}$
\item The norm of the perturbation is bounded, i.e., $|\Delta_j|_2\leq 4\sigma/\sqrt\delta$.
\end{itemize}
Intuitively, the adversary can move a $1-\delta$ fraction of the  data points in each cluster an arbitrary amount towards the convex hull of the means of the remaining clusters. For the remaining $\delta n$, the perturbation should have norm at most $O(\sigma/\sqrt\delta)$. The goal is to still learn the means $\MM_{*,\ell}$ approximately. \cite{BK20} shows that the aforementioned model satisfies Well-Separateness, Proximate Latent Points and Spectrally Bounded Perturbations assumptions. The proof for the Significant Singular Values assumption follows from Lemma \ref{lem:lda_as_simplex}. We note that there has been a flurry of recent progress on adversarial clustering in the strong contamination model, where the input data points are sampled from a mixture of Gaussians distribution and the adversary can corrupt a small fraction of the samples arbitrarily \cite{diakonikolas2018list,hopkins2018mixture,kothari2018robust,diakonikolas2020robustly, bakshi2020outlier}. In our setting, there is no distribution assumption on the data points but the adversary is constrained as the norm of the perturbation is bounded.

\section{Preliminaries}

We use $n,d,$ and $k$ to denote the number of data points, the number of dimensions of the space and the number of vertices
of $\cK$ respectively. 
%Also, we reserve $i,i',i_1,i_2$ to index elements of $[d]$, $j,j',j_1,j_2$ to index $[n]$ and
%$\ell,\ell',\ell_1,\ell_2$ to index $[k]$. $\bA,\bM,\bP$ are reserved for the roles described above.
We use the notation $\AA_{*,j}$ to denote the $j$-th column of matrix $\bA$. 
%For a vector valued random variable $X\in {\bf R}^d$, $\Var(X)$ denotes the covariance matrix of $X$. 
For $\AA\in\mathbb{R}^{d\times n}$ with rank $r$, its  singular value decomposition, denoted by $\texttt{SVD}(\AA) = \UU 
\mathbf{\Sigma} \VV^T$, guarantees that $\UU$ is a $d \times r$ matrix with orthonormal columns, $\VV^T$ is an $r \times n$ matrix with orthonormal rows and $\mathbf{\Sigma}$ is an $r \times r$ diagonal matrix. 
The diagonal entries of $\mathbf{\Sigma}$ are the singular values of $\AA$, denoted by $\sigma_1\ge\sigma_2\ge\ldots\ge\sigma_r$. 
Given an integer $k \leq r$, we define the truncated singular value decomposition of $\AA$ that zeros out
all but the top $k$ singular values of $\AA$, i.e.,  $\AA_k = \UU 
\mathbf{\Sigma}_k \VV^T$, where $\mathbf{\Sigma}_k$ has only $k$ non-zero
entries along the diagonal. It is well-known that the truncated SVD 
computes the best rank-$k$ approximation to $\AA$ under the Frobenius 
norm, i.e., $\AA_k = \min_{\textrm{rank}(\XX)\leq k} \| \AA - \XX\|_F$.
Given an orthonormal basis $\UU$ for a subspace, we use $\PP_{\UU} = \UU \UU^T$ to denote the projection matrix corresponding to the subspace. We consider the following notion of subspace distance:

\begin{definition}[$\sin\Theta$ Distance]
For any two subspaces $\RR$, $\SS$ of $\mathbb{R}^d$, the $\sin\Theta$ distance between $\RR$ and $\SS$ is defined as 
% $\sin\Theta(\RR,\SS) =\underset{u\in\RR}{\max}\,\underset{v\in\SS}{\min}\sin\theta(u,v)
%         =\underset{u\in\RR,|u|=1}{\max}\underset{v\in\SS}{\min}\|u-v\|$.
\begin{equation*}
    \begin{split}
        \sin\Theta(\RR,\SS) =\underset{u\in\RR}{\max}\,\underset{v\in\SS}{\min}\sin\theta(u,v)
        =\underset{u\in\RR,|u|=1}{\max}\underset{v\in\SS}{\min}\|u-v\|.
    \end{split}
\end{equation*}
\end{definition}

We use the notion of spectral low-rank approximation to obtain a compact representation of the input and compute matrix-vector products efficiently. 
We also require the notion of mixed spectral-Frobenius low-rank approximation. 
This guarantee is weaker than spectral-low rank approximation but admits faster algorithms and has been recently used in several sublinear time algorithms~\cite{musco2017sublinear,bakshi2020robust}. 

\begin{definition}[Spectral Low-rank Approximation, Spectral-Frobenius Low-rank Approximation]
\label{def:spectral_lra}
Given a matrix $\AA$, an integer $k$ and $\epsilon>0$, a rank-$k$ matrix $\BB$ satisfies a \textit{relative-error} spectral low-rank approximation guarantee if 
$\|\AA  - \BB\|^2_2 \leq (1+\epsilon)\|\AA - \AA_k \|^2_2$. 
$\BB$ satisfies a mixed spectral-Frobenius low-rank approximation guarantee if 
\begin{equation*}
    \|\AA  - \BB\|^2_2 \leq (1+\epsilon)\|\AA - \AA_k \|^2_2 + \frac{\epsilon}{k} \|\AA - \AA_k\|^2_F.
\end{equation*}
\end{definition}

\section{Technical Overview}
In this section, we provide an overview of our algorithmic techniques and discuss the main challenges we overcome to obtain an input-sparsity time algorithm. 
%Prior work of ~\cite{BK20} uses power iteration to approximate the left top-$k$ singular space $\UU_k$ of $\AA$ using a subspace $\widehat{\VV}$ that is $\poly(\alpha/k)$ close in $\sin\Theta$ distance. 
%Each step of the power iteration requires $O(\nnz(\AA) + dk^2)$ time and they repeat this $\log(d)$ times. 
%Next, they pick a random vector $u_1$ in the subspace spanned $\widehat{\VV}$ and compute $u_1 \cdot \AA$. 

\paragraph{Our Techniques.}

The starting point in \cite{BK20} is that the smoothened polytope, obtained by averaging points in the data matrix $\AA$ is itself close to the latent points in the convex hull of $\cK$ in operator norm. This fact is captured by the following lemma: 
\begin{lemma}[Subset Smoothing]
\label{lem:subset_smoothing}
For any $\S \subset [n]$, let $\AA_{\S}$ be a vector obtained by averaging the columns of $\AA$ indexed by $\S$ and define $\PP_{\S}$ similarly. Then for $\|\AA - \PP \|_2 \leq \sigma \sqrt{n}$, we have $\|\AA_{\S} - \PP_{\S} \|_2 \leq\sigma\sqrt{n/|\S|}$. 
\end{lemma}

Our main insight is that we can approximately optimize a linear function on the smoothed polytope by working with a rank-$k$ spectral approximation to $\AA$ instead. Geometrically, this implies that while the smoothed polytope is perhaps $d$-dimensional, projecting it onto the $k$-dimensional space spanned by the top-$k$ singular values of the data matrix $\AA$ suffices to recover the latent $k$-simplex, $\cK$. This is surprising since the data matrix can contain points significantly far from the latent polytope. Further, this approach presents several challenges: we do not have access to the left singular space of $\AA$ and even if we are provided this subspace exactly, it is unclear why it spans a set of points that approximate vertices of $\cK$. Finally, the points obtained by smoothing the projected polytope have no immediate relation to points in the smoothed high-dimensional polytope considered by \cite{BK20}.

We would like to begin by computing a spectral low-rank approximation (Definition \ref{def:spectral_lra}) for $\AA$.
%and compare the spanto the top-$k$ left singular vectors of $\AA$. 
Since a low-rank approximation to $\AA$ can be represented in factored form $\YY \ZZ^T$, where $\YY$ is $d \times k$ and $\ZZ^T$ is $k\times n$, any matrix-vector product of the form $\YY \ZZ^T \cdot x$ only requires $(n+d)k$ time. Thus optimizing a linear function $k$ times over a smoothed low-rank polytope requires only $(n+d)k^2$ time, circumventing the previous bound of $k\cdot \nnz(\AA)$. 
However, the best known algorithm for spectral low-rank approximation (Theorem 1 in \cite{musco2015randomized}) requires $\widetilde{O}(\nnz(\AA)\cdot k/\sqrt{\epsilon})$ time and thus provides no improvement. A natural direction to pursue is then to compute a Frobenius low-rank approximation (which requires $\nnz(\AA)$ time) for $\AA$ and use this as our proxy. However, a Frobenius low-rank approximation is too coarse to obtain a subspace that is close to the top-$k$ singular vectors of $\AA$.

Instead we compute a mixed spectral-Frobenius low-rank approximation (see Definition \ref{def:spectral_lra}) that runs in $O(\nnz(A) + dk^2)$ time, but the resulting error guarantee is weaker. In particular, it incurs an additive $\epsilon \|\AA -\AA_k \|^2_F/k$ term.
Here, we use the assumption we introduced (the Significant Singular Value assumption) to show that  the low-rank matrix obtained from this algorithm also satisfies a \textit{relative-error} spectral low-rank approximation guarantee. The next challenge is that the aforementioned guarantee only bounds the spectral norm of $\AA - \YY\ZZ^T$ in terms of the $(k+1)$-st singular value of $\AA$. This guarantee does not relate how close the subspaces spanned by the columns and rows of the low-rank approximation are to the top-$k$ singular space of $\AA$.

A key technical contribution of our work is thus to prove that the subspaces obtained via spectral low-rank approximation are close to the true left and right top-$k$ singular space in angular ($\sin\Theta$) distance. We note that such a guarantee is crucial to approximately optimize a linear function over $\AA$. Further, this result provides an intriguing connection between spectral low-rank approximation and power iteration. It is well known that power iteration suffices to obtain a subspace that is close to the top-$k$ subspace of a matrix in  $\sin\Theta$ distance, which at first glance appears much stronger than spectral low-rank approximation. However, our work implies that it suffices to compute a spectral low-rank approximation, which provides a succinct representation of the data matrix and can be computed faster than power iteration in several natural settings.

\begin{Frame}[\textbf{Algorithm \ref{alg:input_sparsity_simplex}} : Learning a Latent $k$-Simplex in Input Sparsity Time]
\label{alg:input_sparsity_simplex}
\textbf{Input}: A matrix $\AA \in \mathbb{R}^{d \times n}$, integer $k$, and $\epsilon>0$.
\begin{enumerate}
    \item Using the algorithm from Lemma \ref{lem:input_sparsity_spectral_lra}, compute  rank-$k$ matrices $\YY, \ZZ$  such that $\YY \ZZ^T$ is a spectral low-rank approximation to $\AA$, i.e., $
    \|\AA - \YY\ZZ^T \|^2_2 \leq (1+\epsilon)\|\AA - \AA_k\|^2_2$. 
    \item Let $\S = \{\emptyset\}$. For each $t \in [k]$,
    \begin{enumerate}
        %\item Let $\S$ be the subset of columns in $\AA$ up until iteration $t-1$. 
        \item Let $\UU_t$ be an orthonormal basis for the vectors in $\S$.
        \item Compute the projection matrix $\PP_t = \UU_t \UU_t^T$ that projects onto the row span of $\S$. 
		\item Let $g \sim \mathcal{N}(0, \II_k)$ and let $\bu_t = g\YY^T(\II_d - \PP_t)\YY\ZZ^T$ be a random vector in $\mathbb{R}^{n}$. 
		Compute $\R_t \subset [n]$, a subset of $\delta n$ indices corresponding to the largest coordinates of $\bu_t$ in absolute value.
        %\item Compute the projection matrix $\PP_t$ that projects on to the row span of vectors in $\R$. \item Let $g \sim \mathcal{N}(0, \II)$ and let $u_t = \YY\ZZ^T(\II - \PP_t)\ZZ g$ be a random vector in $\mathbb{R}^{n}$. Compute $\mathcal{I}_t \subset [n]$, a subset of $\delta n$ indices corresponding to the largest coordinates of $u_t$ in absolute value.
        \item Let $\AA_{\R_{t}}$ be the average of the columns of $\AA$ indexed by $\R_t$. 
        %Update $\S = \S\cup\R_{t}$.
        Update $\S = \S\cup\AA_{R_{t}}$.
    \end{enumerate}
\end{enumerate}
\textbf{Output:} The set of vectors $\AA_{{\R_{1}}}, \AA_{\mathcal{R}_2}, \ldots, \AA_{\mathcal{R}_k}$ as our approximation to the vertices of the latent $k$-simplex $\cK$.  
\end{Frame}

In the context of learning the latent simplex, given a spectral low-rank approximation, $\YY\ZZ^T$, we first restrict to the column span of $\YY$, which w.l.o.g. has orthonormal columns, and iteratively generate $k$ vectors in this subspace. 
In the first iteration, we generate a random vector $g\YY^T$ and compute $g\YY^T \YY\ZZ^T$. 
We then consider the largest $\delta n$ indices of $g\YY^T \YY\ZZ^T$. 
While the resulting vector does not have strong provable guarantees, we show that averaging the columns of $\AA$ corresponding to these indices results in a vector, $\AA_{\R_1}$, which intuitively corresponds to efficiently optimizing a linear function over a low-rank approximation to the smoothened polytope, where the smoothened polytope is obtained by averaging over all subsets of $\delta n$ data points. 
Our next contribution is to show that $\AA_{\R_1}$ obtained by the aforementioned algorithmic process is indeed close to a vertex of $\cK$. 

% Given the Proximate Latent Point assumption \eqref{a2}, picking $\S$ to have at least $\delta n$ points provides hope that $\AA_{\S}$ would be close to a vertex of the latent polytope $\cK$. 
% However, computing $u_1 \cdot \AA$ itself requires $\nnz(\AA)$ time. 
% One approach is to use the set $\AA_{\R_1}= \argmax_{\S: |\S| = \delta n} |u_1\cdot \AA_{\S}|$ to average the corresponding columns in $\AA$ and use the resulting vector $\AA_{\R_1}$ as an approximation to some vertex $\MM_{*,1}$. 

To obtain an approximation to the remaining vertices of $\cK$, we consider the following iterative process: in the $t$-th iteration, consider the subspace $\YY^T(\II - \PP_t)$, where $(\II - \PP_t)$ is the projection onto the orthogonal complement of the span of $\AA_{\R_1},\AA_{\R_2} \ldots \AA_{\R_{t-1}}$.  Then generate a random vector $g\YY^T(\II - \PP_t)$, and compute the largest $\delta n$ coordinates of $g\YY^T(\II - \PP_t) \YY \ZZ^T$. Average the corresponding columns of $\AA$ to obtain $\AA_{\R_t}$ and output this vector. We prove that after iterating $k$ times, the vectors $\AA_{\R_1}, \AA_{\R_1},  \ldots \AA_{\R_k}$ approximate all the vertices of the latent simplex $\cK$ within the desired accuracy and running time.

In contrast, prior work of \cite{BK20} uses power iteration to approximate the left top-$k$ singular space $\UU_k$ of $\AA$ using a subspace $\widehat{\VV}$ that is $\poly(\alpha/k)$ close in $\sin\Theta$ distance. 
Each step of the power iteration uses $O(\nnz(\AA) + dk^2)$ time and is repeated $\log(d)$ times. 
Next, they pick a random vector $u_1$ in the subspace spanned $\widehat{\VV}$ and compute $\AA_{\R_1}= \argmax_{\S: |\S| = \delta n} |u_1\cdot \AA_{\S}|$, using the resulting vector as an approximation to some vertex $\MM_{*,1}$. 

They then repeat the above algorithm $k$ times and in the $i$-th iteration, they pick $u_i$ to be a uniformly random direction in the $k -i$ dimensional subspace constructed as follows: let $\widetilde{\VV}_{i-1}$ be an orthonormal basis for $\AA_{\R_1}, \AA_{\R_2}, \ldots ,\AA_{\R_{i-1}}$. 
% Sample $u_i$ from $\widehat{\VV} \cap \textrm{Null}(\widetilde{\VV}_{i-1})$ and set $\AA_{\R_i}= \argmax_{\S: |\S| = \delta n} |u_i\cdot \AA_{\S}|$. 
Intuitively, this corresponds to sampling a random vector from the subspace orthogonal to the set of vertex approximations picked thus far. 
 The resulting $k$ vectors $\AA_{\R_1},\ldots,\AA_{\R_k}$ are the approximation to the vertices of the latent simplex. Since they directly optimize over the smoothened polytope, the correctness analysis is more straightforward.
 
However, each iteration of the algorithm requires optimizing a linear function over the smoothened polytope and in particular requires computing $u_i\cdot \AA$, and thus, the overall running time is dominated by $k\cdot \nnz(\AA)$. 
 Since the latent simplex satisfies the Well-Separateness condition, the inner product with a random direction is maximized by a unique vertex. 
Intuitively, it appears necessary to project away from the set of vectors obtained up to the $i$-th iteration in order to learn new vertices of $\cK$. The inherently iterative nature of the algorithm combined with matrix-vector product lower bounds indicates that the new algorithmic ideas we introduce are in fact necessary.

\section{Full Analysis}
% We first give the proof of Lemma~\ref{lem:lda_as_simplex}. 
% \begin{proof}
% Assumptions \eqref{a2} and \eqref{a3} follow from Lemma 7.1  in  \cite{BK20}. 
% By Claim 8.1 in  \cite{BK20}, $\sigma_k(\AA) \geq c\alpha  \sqrt{\delta/k} \min_{\ell}\MM_{*,\ell}$. 
% Each column of $\AA$ sums to $1$, so $\|\AA \|^2_F = O(n)$ and $\sigma_k(\AA) \geq \alpha \sqrt{ \delta /k}\|\AA \|_F $. 
% Since $\|\AA-\PP\|_2\le\sigma\sqrt{n}$ by definition of $\sigma$, and $\PP$ consists of $n$ point in the convex hull of $k$ points and thus $\sigma_{k+1}(\PP)=0$, we have $\sigma_{k+1}(\AA) \leq \sigma_{k+1}(\PP) + \|\AA - \PP \|_2 \leq \sigma \sqrt{n} \leq \sigma \|\AA\|_F$. 
% Thus if $\sigma \leq \alpha \sqrt{\delta}/\poly(k)$ for a large enough $\poly(k)$, our Significant Singular Values assumption holds.
% \end{proof}

In this section, we analyze Algorithm \ref{alg:input_sparsity_simplex} and show that it outputs a set of $k$ vectors that approximate the vertices of the latent simplex $K$. Formally, the main theorem we prove is as follows:

\vspace{0.1in}
\noindent\textbf{Theorem \ref{thm:input_sparsity_simplex}}\textit{ (Restated.)} Given input data $\AA$ from the Latent Simplex Model, there exists Algorithm \ref{alg:input_sparsity_simplex} that takes $\widetilde{O}\left(\textsf{nnz}(\AA) + (n+d)\poly(k) \right)$ time to output $k$ vectors $\mathcal{R}_1,\ldots,\mathcal{R}_k$ such that upon permuting the columns of $\MM$, for all $\ell\in[k]$,  we have
\[
\|\mathcal{R}_\ell-\MM_{*,\ell}\|_2\le\frac{300k^4}{\alpha}\frac{\sigma}{\sqrt{\delta}},
\]
with probability at least $1-\frac{1}{\Omega(\sqrt{k})}$.
\vspace{0.1in}

We start with a spectral low-rank approximation for $\AA$. 
We then use the right factor as an approximation to $\Sig_k \VV_k^T$ and the left factor as an approximation to $\UU_k$.  

\begin{lemma}(Input-Sparsity Spectral LRA \cite{cohen2015dimensionality, cohenmm17}.)
\label{lem:input_sparsity_spectral_lra}
Given a matrix $\AA \in \mathbb{R}^{d \times n}$, $\epsilon, \delta >0$ and $k \in \mathbb{N}$, there exists an algorithm that outputs matrices $\YY, \ZZ$, such that with probability at least $1-\delta$, $\|\AA - \YY \ZZ^T \|^2_2 \leq (1+\epsilon)\|\AA -\AA_k\|^2_2+ \frac{\epsilon}{k} \| \AA - \AA_{k}\|^2_F$, 
in time $\widetilde{O}\left(\textsf{nnz}(\AA) + (n+d)\poly(k/\epsilon \delta) \right)$.
\end{lemma}

Under the Significant Singular Values condition \eqref{a4}, setting $\epsilon = \phi$ in Lemma \ref{lem:input_sparsity_spectral_lra} implies with probability $99/100$,

\begin{equation}
    \frac{1}{\poly(k)}\sum^n_{i=k+1}\sigma^2_{i} = \frac{1}{\poly(k)}\|\AA - \AA_k \|^2_F \leq \sigma^2_{k+1} = \|\AA - \AA_k \|^2_2
\end{equation}
and thus $\|\AA - \YY\ZZ^T \|^2_2 \leq 2\|\AA - \AA_k \|^2_2$.  Further, the aforementioned lemma implies such a matrix $\YY\ZZ^T$ can be computed in $\widetilde{O}\left(\textsf{nnz}(\AA) + (n+d)\poly(k/\phi) \right)$ time. Thus the Well-Separateness condition immediately implies that the algorithm from Lemma~\ref{lem:input_sparsity_spectral_lra} is a spectral low-rank approximation.

Next, we show that if $\YY\ZZ^T$ is a good rank $k$ spectral approximation to $\AA$, then the subspace spanned by the columns of $\YY$ must be close to the column span of $\UU_k$, the top-$k$ left singular vectors of $\AA$. 
In fact, the subspace $\YY$ obtained via spectral low-rank approximation is a good approximation to the subspace $\UU_k$ in angular distance. 
The appropriate measure of angular distance between subspaces can be formalized as the principal angle between the subspaces and the corresponding $\sin\Theta$ function. 
%Specifically, for any two subspaces $\RR$, $\SS$ of $\mathbb{R}^d$, let 
%\[\sin\Theta(\RR,\SS)=\underset{u\in\RR}{\max}\,\underset{v\in\SS}{\min}\sin\theta(u,v)=\underset{u\in\RR,|u|=1}{\max}\underset{v\in\SS}{\min}\|u-v\|_2.\]
 Wedin \cite{Wedin72} bounded the $\sin\Theta$ between the SVD subspace of a matrix and the SVD subspace of a slight perturbation of the matrix.
\begin{theorem}[Wedin's $\sin\Theta$ theorem  \cite{Wedin72}]
\label{original:sin:theta}
Let $\RR,\SS\in\mathbb{R}^{d\times n}$ and $0<m\le\ell$ be integers. 
Let $\RR_m$ and $\S\sigma_\ell$ denote the subspaces spanned by the top $m$ singular vectors of $\RR$ and top $\ell$ singular vectors of $\SS$, respectively. 
Suppose $\gamma=\sigma_m(\RR)-\sigma_{\ell+1}(\SS)$. 
Then
\[\sin\Theta(\RR_m,\S\sigma_\ell)\le\frac{\|\RR-\SS\|_2}{\gamma}.\]
\end{theorem}
%%%%%%%%%%%%%%%%%%%%%%%%%%%%%%%%%%%%%%%%%%%%%%%%%%%%%%%%%%%%%%%%%%%%%%%%%%%%%%%%%%%%%%%%%%%%%%%%%%%%%%%%%%%%%%%%%%%%%%%%%%%%%%%%%%%%%%%%
%%%%%%%%%%%%%%%%%%%%%%%%%%%%%%%%%%%%%%%%%%%%%%%%%%%%%%%%%%%%%%%%%%%%%%%%%%%%%%%%%%%%%%%%%%%%%%%%%%%%%%%%%%%%%%%%%%%%%%%%%%%%%%%%%%%%%%%%
%%%%%%%%%%%%%%%%%%%%%%%%%%%%%%%%%%%%%%%%%%%%%%%%%%%%%%%%%%%%%%%%%%%%%%%%%%%%%%%%%%%%%%%%%%%%%%%%%%%%%%%%%%%%%%%%%%%%%%%%%%%%%%%%%%%%%%%%
%%%%%%%%%%%%%%%%%%%%%%%%%%%%%%%%%%%%%%%%%%%%%%%%%%%%%%%%%%%%%%%%%%%%%%%%%%%%%%%%%%%%%%%%%%%%%%%%%%%%%%%%%%%%%%%%%%%%%%%%%%%%%%%%%%%%%%%%
Bhattacharyya and Kannan \cite{BK20} use Wedin's $\sin\Theta$ theorem to measure the distance between the subspace $\UU_k$ spanned by the top $k$ left singular vectors of $\AA$ and the subspace returned by their iterative subspace power method. 
Since we create the sketch $\YY$ for $\UU_k$, we would instead like to argue that $\YY$ and $\UU_k$ are close in $\sin\Theta$ distance.

\begin{lemma}[Proximity of Subspace Projections]
\label{lem:sin:theta}
Let $\YY$ be defined as in Algorithm \ref{alg:input_sparsity_simplex} and let $\UU_k$ be the subspace spanned by the top $k $ left singular vectors of $\AA$.
Let $\PP_{\YY}$ and $\PP_{\UU_k}$ be the $d\times d$ projection matrices onto the row span of $\YY$ and $\UU_k$. 
Then $\|\PP_{\YY}-\PP_{\UU_k}\|_2\le\frac{1}{1000k^{10}}$. 
\end{lemma}
\begin{proof}
Suppose by way of contradiction that $\|\PP_{\YY}-\PP_{\UU_k}\|_2\ge\frac{1}{1000k^{10}}$. 
Note that since $\YY$ and $\UU_k$ are each orthonormal matrices with rank $k$, then 
\[
\|\UU_k\UU_k^T-\YY\YY^T\|_F^2\ge \|\UU_k\UU_k^T-\YY\YY^T\|_2^2 \geq \frac{1}{(1000k^{10})^2}\]
so that 
\begin{equation*}
    \begin{split}
        \|\UU_k\UU_k^T-\YY\YY^T\|_F^2& %=\|\UU_k\UU_k^T\|_F^2+\|\YY\YY^T\|_F^2-2\|\UU_k\YY^T\|_F^2\\
        =\|\UU_k\|_F^2+\|\YY\|_F^2-2\|\UU_k\YY^T\|_F^2\\
        & =2k-2\|\UU_k\YY^T\|_F^2\ge\frac{1}{(1000k^{10})^2}
    \end{split}
\end{equation*}
Hence, $\|\UU_k\YY^T\|_F^2\le k-\frac{1}{(1000k^{10})^2}$. 
Now we would like to show for the sake of contradiction that $\|\AA-\PP_{\YY}\AA\|_2$ is large. 
Thus, for the singular value decomposition $\AA=\UU\Sigma\VV^T$, we write 
\begin{equation*}
    \begin{split}
        \|\AA-\PP_{\YY}\AA\|_2&=\|\UU^T\Sigma-\YY\YY^T\UU^T\Sigma\|_2\\
        & \ge\|\UU_k\UU^T\Sigma-\UU_k\YY\YY^T\UU^T\Sigma\|_2
    \end{split}
\end{equation*}
since $\|\UU_k\|_2\le\|\UU\|_2\le 1$. 
Thus, there exist matrices $\CC_1,\CC_2$ such that
\[\UU_k\UU^T\Sigma-\UU_k\YY\YY^T\UU^T\Sigma=
\begin{bmatrix}
\CC_1 & \CC_2
\end{bmatrix}
\begin{bmatrix}
\Sigma_k & 0\\
0 & \Sigma_{n-k}\\
\end{bmatrix}
,
\]
where $\Sigma_k$ is the diagonal matrix consisting of the top $k$ singular values of $\AA$ and $\Sigma_{n-k}$ is the diagonal matrix consisting of the bottom $n-k$ singular values of $\AA$. 
Now we know that one of the top $k$ eigenvalues of $\UU_k^T\YY\YY^T\UU_k$ is at most $1-\frac{1}{(1000k^{10})^2}$. 
Thus, one of the top $k$ eigenvalues of $\II_k-\CC_1$ is at least $\frac{1}{(1000k^{10})^2}$. 
In particular, let $\lambda$ be such an eigenvalue and let $\bx$ be the corresponding unit eigenvector of $\II-\CC_1$. 
Then we have
%\begin{equation*}
%    \begin{split}
%        \|\UU_k\UU^T\Sigma-\UU_k\YY\YY^T\UU^T\Sigma\|_2& \ge\|\Sigma_k(\II-\CC_1)\bx\|_2 \\
%        & =\sigma_k(\AA)\lambda\\
%        & \ge\frac{1}{(1000k^{10})^2}\sigma_k(\AA).
%    \end{split}
%\end{equation*}
\begin{align*}
\|\UU_k\UU^T\Sigma-\UU_k\YY\YY^T\UU^T\Sigma\|_2& \ge\|(\II-\CC_1)\Sigma_k\bx\|_2\ge\sigma_k(\AA)\lambda\ge\frac{1}{(1000k^{10})^2}\sigma_k(\AA).
\end{align*}
Since the Significant Singular Values assumption implies that $\frac{1}{(1000k^{10})^2}\sigma_k(\AA)>(1+\epsilon)\sigma_{k+1}(\AA)$, this implies that $\|\AA-\PP_{\YY}\AA\|_2>(1+\epsilon)\sigma_{k+1}(\AA)$, which contradicts the assumption that $\YY$ is a good low-rank approximation to $\AA$. 
Thus we have $\|\PP_{\YY}-\PP_{\UU_k}\|_2\le\frac{1}{1000k^{10}}$, as desired. 
\end{proof}
%%%%%%%%%%%%%%%%%%%%%%%%%%%%%%%%%%%%%%%%%%%%%%%%%%%%%%%%%%%%%%%%%%%%%%%%%%%%%%%%%%%%%%%%%%%%%%%%%%%%%%%%%%%%%%%%%%%%%%%%%%%%%%%%%%%%%%%%
%%%%%%%%%%%%%%%%%%%%%%%%%%%%%%%%%%%%%%%%%%%%%%%%%%%%%%%%%%%%%%%%%%%%%%%%%%%%%%%%%%%%%%%%%%%%%%%%%%%%%%%%%%%%%%%%%%%%%%%%%%%%%%%%%%%%%%%%
%%%%%%%%%%%%%%%%%%%%%%%%%%%%%%%%%%%%%%%%%%%%%%%%%%%%%%%%%%%%%%%%%%%%%%%%%%%%%%%%%%%%%%%%%%%%%%%%%%%%%%%%%%%%%%%%%%%%%%%%%%%%%%%%%%%%%%%%
%%%%%%%%%%%%%%%%%%%%%%%%%%%%%%%%%%%%%%%%%%%%%%%%%%%%%%%%%%%%%%%%%%%%%%%%%%%%%%%%%%%%%%%%%%%%%%%%%%%%%%%%%%%%%%%%%%%%%%%%%%%%%%%%%%%%%%%%

Our analysis proceeds via induction on the number of iterations performed by the algorithm. 
Suppose our algorithm has selected $t$ points from our approximation of the top $k$ subspace and these points are reasonably close to $i$ points of the $k$-simplex. In the $(t+1)$-st iteration, 
we again bound the $\sin\Theta$ distance between $\YY^T(\II-\PP_t)$, which corresponds to our approximation of the top $k$ subspace projected away from the selected vectors, and the actual $k$-simplex projected away from the corresponding points closest to our selected vectors. 
This argues that we can continue selecting random vectors in the subspace spanned by $\YY^T(\II-\PP_t)$ as a close approximation to random vectors in $\MM(\II-\PP_t)$.

We first bound the $k$-th singular values of the simplex vertices ($\MM$) and latent variables ($\PP$), leveraging the Well-Separateness and Spectrally Bounded Perturbations assumptions. 
\begin{lemma}[Claim 8.1 in \cite{BK20}]
\label{lem:singular:big}
If the underlying points $\MM$ follow the Well-Separateness and Spectrally Bounded Perturbation assumptions, then
\[\sigma_k(\MM)\ge\frac{1000k^{8.5}}{\alpha^2}\frac{\sigma}{\sqrt{\delta}},\qquad \sigma_k(\PP)\ge\frac{995k^{8.5}\sqrt{n}}{\alpha^2}\sigma.\]
\end{lemma}
We can then upper bound $\sin\Theta$ distance between $\YY$ and $\UU_k$ as follows:
\begin{corollary}
\label{cor:proxy:subspace}
Let $\YY$ be defined as in Algorithm \ref{alg:input_sparsity_simplex} and let $\UU_k$ be the subspace spanned by the top $k$ left singular vectors of $\AA$. 
Then $\sin\Theta(\YY,\UU_k)\le\frac{1}{1000k^{10}}$. 
\end{corollary}
\begin{proof}
By setting $m=k=\ell$ in Theorem~\ref{original:sin:theta}, we have 
\begin{equation*}
    \begin{split}
        \sin\Theta(\YY,\UU_k)&=\sin\Theta(\PP_{\YY},\PP_{\UU_k})\\
        &\le\frac{\|\PP_{\YY}-\PP_{\UU_k}\|_2}{\sigma_k(\YY)-\sigma_{k+1}(\UU_k)}.
    \end{split}
\end{equation*}
\[\]
By definition of $\sigma$, we have that $\|\AA-\PP\|_2\le\sigma\sqrt{n}$. 
Thus, Lemma~\ref{lem:singular:big} implies that $\sigma_k(\AA)\gg1$. 
Since $\YY$ has rank $k$, we have $\sigma_{k+1}(\YY)=0$. 
By Lemma~\ref{lem:sin:theta}, $\sin\Theta(\YY,\UU_k)\le\|\PP_{\YY}-\PP_{\UU_k}\|_2\le\frac{1}{1000k^{10}}$. 
\end{proof}
%%%%%%%%%%%%%%%%%%%%%%%%%%%%%%%%%%%%%%%%%%%%%%%%%%%%%%%%%%%%%%%%%%%%%%%%%%%%%%%%%%%%%%%%%%%%%%%%%%%%%%%%%%%%%%%%%%%%%%%%%%%%%%%%%%%%%%%%
%%%%%%%%%%%%%%%%%%%%%%%%%%%%%%%%%%%%%%%%%%%%%%%%%%%%%%%%%%%%%%%%%%%%%%%%%%%%%%%%%%%%%%%%%%%%%%%%%%%%%%%%%%%%%%%%%%%%%%%%%%%%%%%%%%%%%%%%
%%%%%%%%%%%%%%%%%%%%%%%%%%%%%%%%%%%%%%%%%%%%%%%%%%%%%%%%%%%%%%%%%%%%%%%%%%%%%%%%%%%%%%%%%%%%%%%%%%%%%%%%%%%%%%%%%%%%%%%%%%%%%%%%%%%%%%%%
%%%%%%%%%%%%%%%%%%%%%%%%%%%%%%%%%%%%%%%%%%%%%%%%%%%%%%%%%%%%%%%%%%%%%%%%%%%%%%%%%%%%%%%%%%%%%%%%%%%%%%%%%%%%%%%%%%%%%%%%%%%%%%%%%%%%%%%%
They also showed that vectors in $\UU_k$ are close to the subspace $\MM$:
\begin{lemma}
 \cite{BK20}
\label{bk20:m:vector}
Let $\UU_k$ be the subspace spanned by the top $k$ left singular vectors of $\AA$ and let $\RR$ be any $k$-dimensional subspace of $\mathbb{R}^d$ with $\sin\Theta(\UU_k,\RR)\le\frac{\alpha^2}{1001k^9}$.
Let $\MM$ be the underlying latent $k$-simplex. 
Then for each unit vector $\bx\in\RR$, there exists a vector $\by\in\Span{\MM}$ with $\|\bx-\by\|_2\le\frac{\alpha^2}{500k^{8.5}}$.
\end{lemma}
Since we have $\sin\Theta(\YY,\UU_k)\le\frac{1}{1000k^{10}}$ from Corollary~\ref{cor:proxy:subspace}, then it follows from Lemma~\ref{bk20:m:vector} and the triangle inequality of $\sin\Theta$ distance that vectors in $\YY_k$ are close to the subspace $\MM$:
\begin{corollary}
\label{updated:m:vector}
Let $\YY$ be defined as in Algorithm \ref{alg:input_sparsity_simplex} and let $\RR$ be any $k$-dimensional subspace of $\mathbb{R}^d$ with
\[\sin\Theta(\YY,\RR)\le\frac{\alpha^2}{1000k^9}.\]
Let $\MM$ be the underlying latent $k$-simplex. 
Then for each unit vector $\bx\in\RR$, there exists a vector $\by\in\Span{\MM}$ with $\|\bx-\by\|_2\le\frac{\alpha^2}{500k^{8.5}}$.
\end{corollary}
%%%%%%%%%%%%%%%%%%%%%%%%%%%%%%%%%%%%%%%%%%%%%%%%%%%%%%%%%%%%%%%%%%%%%%%%%%%%%%%%%%%%%%%%%%%%%%%%%%%%%%%%%%%%%%%%%%%%%%%%%%%%%%%%%%%%%%%%
%%%%%%%%%%%%%%%%%%%%%%%%%%%%%%%%%%%%%%%%%%%%%%%%%%%%%%%%%%%%%%%%%%%%%%%%%%%%%%%%%%%%%%%%%%%%%%%%%%%%%%%%%%%%%%%%%%%%%%%%%%%%%%%%%%%%%%%%
%%%%%%%%%%%%%%%%%%%%%%%%%%%%%%%%%%%%%%%%%%%%%%%%%%%%%%%%%%%%%%%%%%%%%%%%%%%%%%%%%%%%%%%%%%%%%%%%%%%%%%%%%%%%%%%%%%%%%%%%%%%%%%%%%%%%%%%%
%%%%%%%%%%%%%%%%%%%%%%%%%%%%%%%%%%%%%%%%%%%%%%%%%%%%%%%%%%%%%%%%%%%%%%%%%%%%%%%%%%%%%%%%%%%%%%%%%%%%%%%%%%%%%%%%%%%%%%%%%%%%%%%%%%%%%%%%

We  then use the following structural result between the first $r$ points selected by Algorithm \ref{alg:input_sparsity_simplex} and the closest $r$ points in the latent $k$-simplex $\MM$. 
\begin{lemma}[Equation 10.21 in \cite{BK20}]
\label{lem:cs}
For  $r \in [k]$ let $\mathcal{R}_1,\ldots,\mathcal{R}_k\in\mathbb{R}^d$ be points such that there exist distinct $\ell_1,\ldots,\ell_r\subseteq[n]$ with
\[\|\mathcal{R}_i-\MM_{*,\ell_i}\|_2\le\frac{300k^4}{\alpha}\frac{\sigma}{\sqrt{\delta}}\]
%for each $i\in[r]$. 
Let $\widehat{\AA}=\mathcal{R}_1\circ\ldots\circ\mathcal{R}_t$ and $\widehat{\MM}=\MM_{*,\ell_1}\circ\ldots\circ\MM_{*,\ell_r}$. 
Then 
\[\|\widehat{\MM}-\widehat{\AA}\|_2\le\frac{k^{4.5}}{\alpha}\frac{\sigma}{\sqrt{\delta}}.\]
\end{lemma}
\begin{proof}
Note that the claim follows immediately from the hypothesis and applying the Cauchy-Schwarz inequality. 
\end{proof}

%%%%%%%%%%%%%%%%%%%%%%%%%%%%%%%%%%%%%%%%%%%%%%%%%%%%%%%%%%%%%%%%%%%%%%%%%%%%%%%%%%%%%%%%%%%%%%%%%%%%%%%%%%%%%%%%%%%%%%%%%%%%%%%%%%%%%%%%
%%%%%%%%%%%%%%%%%%%%%%%%%%%%%%%%%%%%%%%%%%%%%%%%%%%%%%%%%%%%%%%%%%%%%%%%%%%%%%%%%%%%%%%%%%%%%%%%%%%%%%%%%%%%%%%%%%%%%%%%%%%%%%%%%%%%%%%%
%%%%%%%%%%%%%%%%%%%%%%%%%%%%%%%%%%%%%%%%%%%%%%%%%%%%%%%%%%%%%%%%%%%%%%%%%%%%%%%%%%%%%%%%%%%%%%%%%%%%%%%%%%%%%%%%%%%%%%%%%%%%%%%%%%%%%%%%
%%%%%%%%%%%%%%%%%%%%%%%%%%%%%%%%%%%%%%%%%%%%%%%%%%%%%%%%%%%%%%%%%%%%%%%%%%%%%%%%%%%%%%%%%%%%%%%%%%%%%%%%%%%%%%%%%%%%%%%%%%%%%%%%%%%%%%%%
We first bound the $\sin\Theta$ distance between $\Span{\MM}\cap\Null(\widehat{\MM})$ and $\YY(\II_d-\PP_r)$. 
This essentially says that we can work in the subspace $\YY(\II_d-\PP_r)$ rather than $\Span{\MM}\cap\Null(\widehat{\MM})$ and we will not incur too much error. 

Next, we prove our lemma relating angular distance of the subspace obtained in the $i$-th iteration of the algorithm ($\YY(\II - \PP_i)$) to the optimal subspace ($\MM(\II - \PP_i)$).  

\begin{lemma}[Angular Distance between Subspaces.]
\label{bk:ten:one}
For some $r \in [k]$, let $\widehat{\MM}=\MM_{*,\ell_1}\circ\ldots\circ\MM_{*,\ell_r}$ be the matrix with $r$ columns corresponding to vertices of the latent $k$-simplex $\MM$ closest to the first $r$ points selected by Algorithm \ref{alg:input_sparsity_simplex}, $\AA_{\R_1},\ldots,\AA_{\R_{r}}$, respectively. 
Suppose $\|\AA_{\R_{i}}-\MM_{*,\ell_i}\|_2\le\frac{300k^4}{\alpha}\frac{\sigma}{\sqrt{\delta}}$ 
for each $i\in[r]$. 
Let $\PP_r$ be the projection matrix orthogonal to $\AA_{\R_1},\ldots,\AA_{\R_r}$. 
Then, 
\begin{align*}
\sin\Theta\left(\YY(\II_d-\PP_r),\Span{\MM}\cap\Null(\widehat{\MM})\right)&\le\frac{\alpha}{100k^4}\\ \sin\Theta\left(\Span{\MM}\cap\Null(\widehat{\MM}),\YY(\II_d-\PP_r)\right)&\le\frac{\alpha}{100k^4}. 
\end{align*}
%Then, with probability at least $1-\Omega(1/\sqrt{k})$, $\sin\Theta\left(\YY(\II_d-\PP_r),\Span{\MM}\cap\Null(\widehat{\MM})\right)\le\alpha/{100k^4}$ and $\sin\Theta\left(\Span{\MM}\cap\Null(\widehat{\MM}),\YY(\II_d-\PP_r)\right)\le\alpha/{100k^4}$.
\end{lemma}

% \vspace{0.1in}
% \noindent\textbf{Lemma~\ref{bk:ten:one}}\textit{ (Angular Distance between Subspaces, Restated.)
% Let $\widehat{\MM}=\MM_{*,\ell_1}\circ\ldots\circ\MM_{*,\ell_r}$ be the $r$ points in the latent $k$-simplex $\MM$ closest to $\R_1,\ldots,\R_{r}$, the first $r$ points selected by our algorithm, respectively.
% Suppose $\|\R_{i}-\MM_{*,\ell_i}\|_2\le\frac{300k^4}{\alpha}\frac{\sigma}{\sqrt{\delta}}$ for each $i\in[r]$. 
% Let $\PP_r$ be the projection matrix orthogonal to $\R_1,\ldots,\R_{r}$. 
% Then 
% \begin{align*}
% \sin\Theta\left(\YY(\II_d-\PP_r),\Span{\MM}\cap\Null(\widehat{\MM})\right)&\le\frac{\alpha}{100k^4}\\ \sin\Theta\left(\Span{\MM}\cap\Null(\widehat{\MM}),\YY(\II_d-\PP_r)\right)&\le\frac{\alpha}{100k^4}. 
% \end{align*}
% }
% \vspace{0.1in}
\begin{proof}
Let $\by\in\YY(\II_d-\PP_r)$ be a unit vector. 
By Corollary~\ref{updated:m:vector}, there exists $\bx\in\Span{\MM}$ with 
\begin{equation}
\label{eqn:bx:by}
\|\bx-\by\|_2\le\frac{\alpha^2}{500k^{8.5}}.
\end{equation} 
Let $\bz=\bx-\widehat{\MM}\widehat{\MM}^\dagger\bx$ be the component of $\bx$ in $\Null(\widehat{\MM})$. 
Note that $\widehat{\MM}\widehat{\MM}^\dagger$ is a projection matrix and thus $\|\widehat{\MM}\widehat{\MM}^\dagger\|_2\le 1$. 
Then we have
\begin{align*}
\|\bx-\bz\|_2&\le\|\widehat{\MM}\widehat{\MM}^\dagger(\bx-\by)\|_2+\|\widehat{\MM}\widehat{\MM}^\dagger\by\|_2\\
&\le\|\bx-\by\|_2+\|\widehat{\MM}(\widehat{\MM}^T\widehat{\MM})^{-1}(\widehat{\MM}^T-\widehat{\AA}^T)\by\|_2
\end{align*}
where $\widehat{\AA}=\R_1\circ\ldots\circ\mathcal{R}_t$ so that $\widehat{\AA}^T\by=0$ since $\PP_r$ projects away from $\widehat{\AA}$. 
We also have $\|\widehat{\MM}(\widehat{\MM}^T\widehat{\MM})^{-1}\|_2=\frac{1}{\sigma_r(\widehat{\MM})}$. 
Thus by (\ref{eqn:bx:by}) and Lemma~\ref{lem:cs}, we have
\begin{equation*}
    \begin{split}
        \|\bx-\bz\|_2&\le\|\bx-\by\|_2+\frac{1}{\sigma_r(\widehat{\MM})}\|(\widehat{\MM}^T-\widehat{\AA}^T)\by\|_2 \\
        &\le\frac{\alpha^2}{500k^{8.5}}+\frac{k^{4.5}\sigma}{\alpha\sqrt{\delta}\sigma_k(\widehat{\MM})}.
    \end{split}
\end{equation*}
Hence by the triangle inequality and Lemma~\ref{lem:singular:big}, we have $\|\by-\bz\|_2\le\frac{\alpha}{100k^4}$. 
Since $\by\in\YY(\II_d-\PP_r)$ and $\bz\in\Span{\MM}\cap\Null(\widehat{\MM})$, then by definition of the $\sin\Theta$ distance, it follows that 
\begin{equation*}
\sin\Theta\left(\YY(\II_d-\PP_r),\Span{\MM}\cap\Null(\widehat{\MM})\right)\le\frac{\alpha}{100k^4},
\end{equation*} 
proving the first part of the claim. 

To prove the second half of the claim, it suffices to show that the dimension of $\YY(\II_d-\PP_r)$ is $k-r$, since $\Span{\MM}\cap\Null(\widehat{\MM})$ has dimension $k-r$ and the $\sin\Theta$ distance is symmetric between two subspaces of the same dimension. 
By construction, $\YY$ has dimension $k$ so that $\YY(\II_d-\PP_r)$ has dimension at least $k-r$. 
But if $\YY(\II_d-\PP_r)$ has dimension larger than $k-r$, then there exists a set of orthonormal vectors $\bu_1,\ldots,\bu_{k-r+1}\in\YY(\II_d-\PP_r)$. 
By the first part of the claim and the definition of the $\sin\Theta$ distance, there exists a set of corresponding vectors $\bv_1,\ldots,\bv_{k-r+1}\in\Span{\MM}\cap\Null(\widehat{\MM})$ such that $\|\bu_i-\bv_j\|_2<\frac{\alpha}{100k^4}$. 
But then for $a\neq b$, we have by the triangle inequality and the fact that $\bu_a\cdot\bu_b=0$,
\begin{equation*}
    \begin{split}
        |\bv_a\cdot\bv_b|& \le|\bu_a\cdot\bu_b|+|(\bv_a-\bu_a)\cdot\bu_b|+|\bv_a\cdot(\bv_b-\bu_b)|\\
        &\le\frac{\alpha}{50k^4}
    \end{split}
\end{equation*}
Similarly, since $\bu_a\cdot\bu_a=1$, we have
\begin{equation*}
    \begin{split}
        |\bv_a\cdot\bv_a|& \ge|\bu_a\cdot\bu_a|-|(\bv_a-\bu_a)\cdot\bu_a|-|\bv_a\cdot(\bv_a-\bu_a)|\\
        &\ge 1-\frac{\alpha}{50k^4}.
    \end{split}
\end{equation*}
Thus if $\VV=\bv_1\circ\ldots\circ\bv_{k-r+1}\in\mathbb{R}^{d\times k-r+1}$ is formed by concatenating the vectors $\bv_1,\ldots,\bv_{k-r+1}$, then $\VV^T\VV$ is diagonally-dominant. 
Hence, $\VV^T\VV$ is nonsingular, so $\bv_1,\ldots,\bv_{k-r+1}$ must be linearly independent vectors in $\Span{\MM}\cap\Null(\widehat{\MM})$, which contradicts the fact that its dimension is $k-r$. 
Therefore, the dimension of $\YY(\II_d-\PP_r)$ must be $k-r$, and so $\sin\Theta\left(\Span{\MM}\cap\Null(\widehat{\MM}),\YY(\II_d-\PP_r)\right)\le\frac{\alpha}{100k^4}$. 
\end{proof}
We now recall a structural lemma from  \cite{BK20}.
\begin{lemma}[Claim 10.1 in  \cite{BK20}]
\label{structural:choices}
Let $a,b\notin\{\ell_1,\ldots,\ell_r\}$ be distinct indices. 
Then
\[\|\Proj(\MM_{*,a}-\MM_{*,b},\Null(\widehat{\MM})\|_2\ge\alpha\max_{\ell}\|\MM_{*,\ell}\|_2.\]
\end{lemma}
%%%%%%%%%%%%%%%%%%%%%%%%%%%%%%%%%%%%%%%%%%%%%%%%%%%%%%%%%%%%%%%%%%%%%%%%%%%%%%%%%%%%%%%%%%%%%%%%%%%%%%%%%%%%%%%%%%%%%%%%%%%%%%%%%%%%%%%%
%%%%%%%%%%%%%%%%%%%%%%%%%%%%%%%%%%%%%%%%%%%%%%%%%%%%%%%%%%%%%%%%%%%%%%%%%%%%%%%%%%%%%%%%%%%%%%%%%%%%%%%%%%%%%%%%%%%%%%%%%%%%%%%%%%%%%%%%
%%%%%%%%%%%%%%%%%%%%%%%%%%%%%%%%%%%%%%%%%%%%%%%%%%%%%%%%%%%%%%%%%%%%%%%%%%%%%%%%%%%%%%%%%%%%%%%%%%%%%%%%%%%%%%%%%%%%%%%%%%%%%%%%%%%%%%%%
%%%%%%%%%%%%%%%%%%%%%%%%%%%%%%%%%%%%%%%%%%%%%%%%%%%%%%%%%%%%%%%%%%%%%%%%%%%%%%%%%%%%%%%%%%%%%%%%%%%%%%%%%%%%%%%%%%%%%%%%%%%%%%%%%%%%%%%%

Now we need to show that our algorithm is (1) well-defined and (2) preserves the invariant that the $(i+1)$-st point sampled from $\YY^T(\II-\PP_i)$ will also be reasonably close to some different point of the $k$-simplex. 
We show the selected procedure is well-defined in Lemma~\ref{bk:ten:two} by arguing that there exists a unique solution to the maximization problem. 

\begin{lemma}[Optimization is Well-Defined]
\label{bk:ten:two}
Let $\bu\in\mathbb{R}^d$ be a random unit vector in the space of $\YY^T(\II_d - \PP_r)$, where $\PP_r$ is the orthogonal projection to $\AA_{\mathcal{R}_1},\ldots,\AA_{\mathcal{R}_r}$. 
%Let $\bu=g\YY^T(\II_d - \PP_r)\in\mathbb{R}^d$, where $\PP_r$ is the orthogonal projection to $\AA_{\R_1},\ldots,\AA_{\R_r}$. 
Then there exists a constant $c>0$ so that with probability at least $1-c/{k^{1.5}}$:
\begin{enumerate} 
\item
For all distinct $a,b\notin\{\ell_1,\ldots,\ell_r\}$, then $|\bu\cdot(\MM_{*,a}-\MM_{*,b})|\ge\frac{0.097}{k^4}\alpha\max_{\ell}\|\MM_{*,\ell}\|_2$.
\item
For all $a\notin\{\ell_1,\ldots,\ell_r\}$, then $|\bu\cdot\MM_{*,a}|\ge\frac{0.0989}{k^4}\alpha\max_{\ell}\|\MM_{*,\ell}\|_2$.
\end{enumerate}
\end{lemma}

% We now show that if $\bu=g\YY^T(\II_d - \PP_r)\in\mathbb{R}^d$, then $|\bu\cdot\bx|$ has a clear optimum over $\bx$ chosen from the $k$ vertices of the simplex $\MM$. 
% This shows that our optimization procedure is well-defined.
% \vskip0.1in
% \noindent\textbf{Lemma~\ref{bk:ten:two}}\textit{ (Optimization is Well-Defined)
% Let $\widehat{\MM}=\MM_{*,\ell_1}\circ\ldots\circ\MM_{*,\ell_r}$ be the $r$ points in the latent $k$-simplex $\MM$ closest to the first $r$ points selected by Algorithm \ref{alg:input_sparsity_simplex}, $\R_1,\ldots,\R_{r}$, respectively. 
% Suppose
% \[\|\mathcal{R}_i-\MM_{*,\ell_i}\|_2\le\frac{300k^4}{\alpha}\frac{\sigma}{\sqrt{\delta}}\] 
% for each $i\in[r]$. 
% Let $\bu\in\mathbb{R}^d$ be a random unit vector in the space of $\YY^T(\II_d - \PP_r)$, where $\PP_r$ is the orthogonal projection to $\mathcal{R}_1,\ldots,\mathcal{R}_r$. 
% %Let $\bu=g\YY^T(\II_d - \PP_r)\in\mathbb{R}^d$, where $\PP_r$ is the orthogonal projection to $\R_1,\ldots,\R_{r}$. 
% Then there exists a constant $c>0$ so that with probability at least $1-\frac{c}{k^{1.5}}$:
% \begin{enumerate}
% \item
% For all distinct $a,b\notin\{\ell_1,\ldots,\ell_r\}$, then $|\bu\cdot(\MM_{*,a}-\MM_{*,b})|\ge\frac{0.097}{k^4}\alpha\max_{\ell}\|\MM_{*,\ell}\|_2$.
% \item
% For all $a\notin\{\ell_1,\ldots,\ell_r\}$, then $|\bu\cdot\MM_{*,a}|\ge\frac{0.0989}{k^4}\alpha\max_{\ell}\|\MM_{*,\ell}\|_2$.
% \end{enumerate}
% }
\begin{proof}
For $a\notin\{\ell_1,\ldots,\ell_r\}$, let $\bq_a$ be the projection of $\MM_{*,a}$ onto $\Null(\widehat{\MM})$ and $\bp_a$ be the projection of $\MM_{*,a}$ onto $\Span{\widehat{\MM}}$. 
By the Well-Separateness assumption, we have $\|\bq_a\|_2\ge\alpha\max_{\ell}\|\MM_{*,\ell}\|_2$.
Let $\bw_a$ be defined so that $\bp_a=\widehat{\MM}\bw_a$. 
Since $\|\bp_a\|_2\le\|\MM_{*,a}\|_2$ and $\sigma_r(\widehat{\MM})\le \sigma_k(\MM)$, then Lemma~\ref{lem:singular:big} gives
\begin{equation}
\label{wa:upperbound}
\|\bw_a\|_2\le\frac{\|\bp_a\|_2}{\sigma_r(\widehat{\MM})}\le\frac{\|\MM_{*,a}\|_2\alpha^2}{1000k^{8.5}}\frac{\sqrt{\delta}}{\sigma}.
\end{equation}
Since $\widehat{\AA}\bu=0$, we can also write
\begin{equation*}
    \begin{split}
        \bu\cdot\MM_{*,a}& =\bu\cdot\bq_a+\bu\cdot\bp_a\\
        & =\bu\cdot\Proj(\bq_a,\YY^T(\II_d - \PP_r))+\bu^T(\widehat{\MM}-\widehat{\AA})\bw_a.
    \end{split}
\end{equation*}
By Lemma~\ref{lem:cs}, (\ref{wa:upperbound}), and normalizing so that $\|\bu\|_2=1$, we have
\begin{equation}
\label{uma:upperbound}
\begin{split}
    |\bu\cdot(\MM_{*,a}-\cdot\Proj(\bq_a,\YY^T(\II_d - \PP_r)))|& \le\|\widehat{\MM}-\widehat{\AA}\|_2\|\bw_a\|_2\\
    & \le\frac{\alpha\|\MM_{*,a}\|_2}{1000k^4}.
\end{split}
\end{equation}
The same holds for $\bu\cdot(\MM_{*,a}-\MM_{*,b})$, so that 
\begin{equation}
\label{uma:umb:upperbound}
\begin{split}
    &|\bu\cdot(\MM_{*,a}-\MM_{*,b})-\bu\cdot\Proj(\bq_a-\bq_b,\YY^T(\II_d - \PP_r))|\\
    & \le\frac{\|\MM_{*,a}-\MM_{*,b}\|_2\alpha}{1000k^4}.
\end{split}
\end{equation}
Let $\mathcal{E}$ be the event that:
\begin{enumerate}
\item
For all $a$, $|\bu\cdot\Proj(\bq_a,\YY^T(\II_d - \PP_r))|\ge\frac{1}{10k^4}\|\Proj(\bq_a,\YY^T(\II_d - \PP_r))\|_2$. 
\item
For all $a\neq b$, $|\bu\cdot\Proj(\bq_a-\bq_b,\YY^T(\II_d - \PP_r))|\ge\frac{1}{10k^4}\|\Proj(\bq_a-\bq_b,\YY^T(\II_d - \PP_r))\|_2$. 
\end{enumerate}
Note that $|\bu\cdot\Proj(\bq_a,\YY^T(\II_d - \PP_r))|\ge\frac{1}{10k^4}\|\Proj(\bq_a,\YY^T(\II_d - \PP_r))\|_2$ holds as long as $\bu\cdot\Proj(\bq_a,\YY^T(\II_d - \PP_r))\neq 0$. 
Since the volume of the set $\{\bx\in\YY^T(\II_d-\PP_r): \bu\cdot\bx=0\}$ is at most $\sqrt{k}$ times the volume of the unit ball $\{\bx\in\YY^T(\II_d-\PP_r):\|\bx\|_2=1\}$, then by taking a union bound over at most $k^2$ indices, it follows that $\mathcal{E}$ holds with probability at least $1-\frac{1}{k^{1.5}}$. 

By Lemma~\ref{bk:ten:one}, there exists $\bq'_a\in\YY^T(\II_d-\PP_r)$ such that $\|\bq'_a-\bq_a\|_2\le\frac{\alpha\|\bq_a\|_2}{100k^4}$. 
Hence for $k\ge 2$, $\|\bq_a-\Proj(\bq_a,\YY^T(\II_d-\PP_r))\|_2\le\frac{\alpha\|\bq_a\|_2}{100k^4}\le\frac{\|\bq_a\|_2}{1600}$. 
This implies $\|\Proj(\bq_a,\YY^T(\II_d-\PP_r))\|_2\ge0.999\|\bq_a\|_2$. 
Then conditioning on $\mathcal{E}$, 
\begin{equation*}
    \begin{split}
        |\bu\cdot\Proj(\bq_a,\YY^T(\II_d-\PP_r))|& \ge\frac{\|\Proj(\bq_a,\YY^T(\II_d-\PP_r))\|_2}{10k^4}\\
        &\ge\frac{0.999\|\bq_a\|_2}{10k^4}\\
        &\ge\frac{0.999\max_{\ell}\|\MM_{*,\ell}\|_2}{10k^4},
    \end{split}
\end{equation*}
where the last inequality follows since $\|\bq_a\|_2\ge\|\Proj(\MM_{*,a},\Null(\MM\setminus\MM_{*,a}))\|_2\ge\alpha\max_{\ell}\MM_{*,\ell}$ by the Well-Separateness assumption. 
Hence by (\ref{uma:upperbound}), it follows that for all $a\notin\{\ell_1,\ldots,\ell_r\}$,
\begin{equation*}
    \begin{split}
        |\bu\cdot\MM_{*,a}|& \ge|\bu*\Proj(\bu,\YY^T(\II_d-\PP_r))-\frac{\alpha\|\MM_{*,a}\|_2}{1000k^4}\\
        &\ge\frac{0.0989\alpha\max_{\ell}\|\MM_{*,\ell}\|_2}{k^4},
    \end{split}
\end{equation*}
which proves the second half of the claim.

To prove the first half of the claim, note that conditioned on $\mathcal{E}$, then (\ref{uma:umb:upperbound}) implies
\begin{equation*}
    \begin{split}
        |\bu\cdot(\MM_{*,a}-\MM_{*,b})|&\ge|\bu\cdot\Proj(\bq_a-\bq_b,\YY^T(\II_d - \PP_r))|\\
        &-\frac{\|\MM_{*,a}-\MM_{*,b}\|_2\alpha}{1000k^4}\\
        &\ge\frac{\|\Proj(\bq_a-\bq_b,\YY^T(\II_d - \PP_r))\|_2}{10k^4}\\
        &-\frac{\|\MM_{*,a}-\MM_{*,b}\|_2\alpha}{1000k^4}.
    \end{split}
\end{equation*}
By Lemma~\ref{bk:ten:one}, there exists $\bv\in\YY^T(\II_d-\PP_r)$ such that $\|\bv-(\bq_a-\bq_b)\|_2\le\frac{\alpha\|\bq_a-\bq_b\|_2}{100k^4}$. 
Thus, $\|\Proj(\bq_a-\bq_b,\YY^T(\II_d-\PP_r))\|_2\ge0.99\|\bq_a\|_2\ge0.99\alpha\max_{\ell}\|\MM_{*,\ell}\|_2$, by Lemma~\ref{structural:choices}. 
Since $\frac{\|\MM_{*,a}-\MM_{*,b}\|_2\alpha}{1000k^4}\le\frac{2\alpha\max_{\ell}\|\MM_{*,\ell}\|_2}{1000k^4}$, it follows that $|\bu\cdot(\MM_{*,a}-\MM_{*,b})|\ge\frac{0.097}{k^4}\alpha\max_{\ell}\|\MM_{*,\ell}\|_2$.
\end{proof}
%%%%%%%%%%%%%%%%%%%%%%%%%%%%%%%%%%%%%%%%%%%%%%%%%%%%%%%%%%%%%%%%%%%%%%%%%%%%%%%%%%%%%%%%%%%%%%%%%%%%%%%%%%%%%%%%%%%%%%%%%%%%%%%%%%%%%%%%
%%%%%%%%%%%%%%%%%%%%%%%%%%%%%%%%%%%%%%%%%%%%%%%%%%%%%%%%%%%%%%%%%%%%%%%%%%%%%%%%%%%%%%%%%%%%%%%%%%%%%%%%%%%%%%%%%%%%%%%%%%%%%%%%%%%%%%%%
%%%%%%%%%%%%%%%%%%%%%%%%%%%%%%%%%%%%%%%%%%%%%%%%%%%%%%%%%%%%%%%%%%%%%%%%%%%%%%%%%%%%%%%%%%%%%%%%%%%%%%%%%%%%%%%%%%%%%%%%%%%%%%%%%%%%%%%%
%%%%%%%%%%%%%%%%%%%%%%%%%%%%%%%%%%%%%%%%%%%%%%%%%%%%%%%%%%%%%%%%%%%%%%%%%%%%%%%%%%%%%%%%%%%%%%%%%%%%%%%%%%%%%%%%%%%%%%%%%%%%%%%%%%%%%%%%

We next show that the selected index is not among the previously selected indices. 
Thus, we obtain a new index at each iteration, which implies that we only need $k$ iterations.
\begin{lemma}
\label{bk:ten:extra}
Let $\widehat{\MM}=\MM_{*,\ell_1}\circ\ldots\circ\MM_{*,\ell_r}$ be the $r$ points in the latent $k$-simplex $\MM$ closest to the first $r$ points selected by Algorithm \ref{alg:input_sparsity_simplex}, $\mathcal{R}_1,\ldots,\mathcal{R}_r$, respectively. 
Suppose
\[\|\mathcal{R}_i-\MM_{*,\ell_i}\|_2\le\frac{300k^4}{\alpha}\frac{\sigma}{\sqrt{\delta}}\] 
for each $i\in[r]$. 
Let $\bu\in\mathbb{R}^d$ be a random unit vector in the space of $\YY^T(\II_d - \PP_r)$, where $\PP_r$ is the orthogonal projection to $\mathcal{R}_1,\ldots,\mathcal{R}_r$. 
Let
\[\ell_{r+1}=\begin{cases}\argmax_{\ell}\bu\cdot\MM_{*,\ell}\qquad\text{if }\bu\cdot\R_{r+1}\ge 0\\
\argmin_{\ell}\bu\cdot\MM_{*,\ell}\qquad\text{if }\bu\cdot\R_{r+1}<0\end{cases}.\] 
Then $\ell_{r+1}\notin\{\ell_1,\ldots,\ell_r\}$. 
\end{lemma}
\begin{proof}
We consider the case $\bu\cdot\R_{r+1}\ge 0$ as the analysis for the case $\bu\cdot\R_{r+1}<0$ is symmetric. 
Let $\ell_{r+1}=\argmax_{\ell}\bu\cdot\MM_{*,\ell}$. 
Suppose by way of contradiction that $\ell_{r+1}\in\{\ell_1,\ldots,\ell_r\}$. 
Without loss of generality, let $\ell_{r+1}=\ell_1$. 
Since $\|\R_1-\MM_{*,\ell_1}\|_2\le\frac{300k^4}{\alpha}\frac{\sigma}{\sqrt{\delta}}$ and $\bu\cdot\R_1$, then
\[\bu\cdot\MM_{*,\ell_i}\|_2\le\bu\cdot\R_{i}+\frac{300k^4}{\alpha}\frac{\sigma}{\sqrt{\delta}}=\frac{300k^4}{\alpha}\frac{\sigma}{\sqrt{\delta}}.\] 
Since $\ell_1=\argmax_{\ell}\bu\cdot\MM_{*,\ell}$, then $\bu\cdot\MM_{*,\ell}\le\bu\cdot\MM_{*,\ell_1}$ for all $\ell$. 
Thus $\bu\cdot\PP_{*,S}\le\frac{300k^4}{\alpha}\frac{\sigma}{\sqrt{\delta}}$ for any set of indices $S\subseteq[n]$ inside the convex hull of $\MM$. 
In conjunction, Lemma~\ref{lem:ap:diff} implies 
\begin{equation}
\label{uar:upper}
\bu\cdot\AA_{*,\R_{r+1}}\le\bu\cdot\PP_{*,\R_{r+1}}+\frac{\sigma}{\sqrt{\delta}}\le\left(\frac{300k^4}{\alpha}+1\right)\frac{\sigma}{\sqrt{\delta}}.
\end{equation}
Recall that by Lemma~\ref{lem:input_sparsity_spectral_lra}, $\|\AA - \YY \ZZ^T \|^2_2 \leq (1+\epsilon)\|\AA -\AA_k\|^2_2+ \frac{\epsilon}{k} \| \AA - \AA_{k}\|^2_F$ and thus $\|\AA-\YY\ZZ^T\|\le(1+2\epsilon)\|\AA-\AA_k\|_2$, given the Significant Singular Values assumption. 
Since $\AA_{*,\R_{r+1}}$ is a subset of $\delta n$ columns of $\AA$ and $\R_{r+1}$ is a subset of $\delta n$ columns of $\YY$, then for $\epsilon<1$,
%\[\|\AA-\YY\|_2\le(1+2\epsilon)\sigma_{k+1}(\AA)\le 2\|\AA-\PP\|_2\le2\sigma
\begin{align*}
\bu\cdot\R_{r+1}&\le\bu\cdot\AA_{*,\R_{r+1}}+\bu\cdot(\R_{r+1}-\AA_{*,\R_{r+1}})\\
&\le\left(\frac{300k^4}{\alpha}+1\right)\frac{\sigma}{\sqrt{\delta}}+\frac{3}{\sqrt{\delta n}}\|\AA-\AA_k\|_2,
\end{align*}
where the last step follows from (\ref{uar:upper}) and applying the Cauchy-Schwarz inequality and the fact that $\bu$ is a unit vector. 
Since $\PP$ has rank $k$ and $\AA_k$ is the best rank $k$ approximation to $\AA$, then $\|\AA-\AA_k\|_2\le\|\AA-\PP\|_2$ so that
\begin{align}
\bu\cdot\R_{r+1}&\le\left(\frac{300k^4}{\alpha}+1\right)\frac{\sigma}{\sqrt{\delta}}+\frac{3}{\sqrt{\delta n}}\|\AA-\PP\|_2,\nonumber\\ 
&\le\left(\frac{300k^4}{\alpha}+1\right)\frac{\sigma}{\sqrt{\delta}}+\frac{3\sigma}{\sqrt{\delta}}\\
&=\left(\frac{300k^4}{\alpha}+4\right)\frac{\sigma}{\sqrt{\delta}},\label{bur:upper}
\end{align}
since $\|\AA-\PP\|_2\le\sigma\sqrt{n}$ by definition of $\sigma$. 
However for $t\notin\{\ell_1,\ldots,\ell_r\}$, Lemma~\ref{lem:ap:diff} and the Proximate Latent Points assumption imply the existence of a set $\sigma_t$ of $\delta n$ columns such that
\begin{equation}
\label{bu:ast:lower}
\begin{split}
    |\bu\cdot\AA_{*,\sigma_t}|&\ge|\bu\cdot\PP_{*,\sigma_t}|-\frac{\sigma}{\sqrt{\delta}}\\
    &\ge|\bu\cdot\MM_{*,t}|-\frac{5\sigma}{\sqrt{\delta}}\\
    &\ge\frac{0.0989}{k^4}\alpha\max_{\ell}\|\MM_{*,\ell}\|_2-\frac{5\sigma}{\sqrt{\delta}},
\end{split}
\end{equation}
where the last step follows from Lemma~\ref{bk:ten:two}. 
Moreover, $\sigma_t$ has $\delta n$ columns, so again by applying the Cauchy-Schwarz inequality and the fact that $\bu$ is a unit vector, we have
\begin{equation}
\label{bu:astyst}
\begin{split}
    |\bu\cdot(\AA_{*,\sigma_t}-\YY_{*,\sigma_t})|&\le\frac{1}{\sqrt{\delta n}}\|\AA-\AA_k\|_2\\
    &\le\frac{1}{\sqrt{\delta n}}\|\AA-\PP\|_2\le\frac{3\sigma}{\sqrt{\delta}}.
\end{split}
\end{equation}
where the last two inequalities come from the fact that $\PP$ has rank $k$ and $\|\AA-\PP\|_2\le\sigma\sqrt{n}$ by definition of $\sigma$. 

Thus from (\ref{bu:ast:lower}) and (\ref{bu:astyst}),
\begin{equation*}
    \begin{split}
        |\bu\cdot\YY_{*,\sigma_t}|&\ge|\bu\cdot\AA_{*,\sigma_t}|-|\bu\cdot(\AA_{*,\sigma_t}-\YY_{*,\sigma_t})|\\
        & \ge\frac{0.0989}{k^4}\alpha\max_{\ell}\|\MM_{*,\ell}\|_2-\frac{8\sigma}{\sqrt{\delta}}.
    \end{split}
\end{equation*}
However by the Spectrally Bounded Perturbation assumption, we have $|\bu\cdot\YY_{*,\sigma_t}|\ge\frac{2400k^5}{\alpha}\frac{\sigma}{\sqrt{\delta}}-\frac{8\sigma}{\sqrt{\delta}}$, which contradicts the maximality of $\R_{r+1}$ in (\ref{bur:upper}). 
Therefore, it holds that $\ell_{r+1}\notin\{\ell_1,\ldots,\ell_r\}$.
\end{proof}
%%%%%%%%%%%%%%%%%%%%%%%%%%%%%%%%%%%%%%%%%%%%%%%%%%%%%%%%%%%%%%%%%%%%%%%%%%%%%%%%%%%%%%%%%%%%%%%%%%%%%%%%%%%%%%%%%%%%%%%%%%%%%%%%%%%%%%%%
%%%%%%%%%%%%%%%%%%%%%%%%%%%%%%%%%%%%%%%%%%%%%%%%%%%%%%%%%%%%%%%%%%%%%%%%%%%%%%%%%%%%%%%%%%%%%%%%%%%%%%%%%%%%%%%%%%%%%%%%%%%%%%%%%%%%%%%%
%%%%%%%%%%%%%%%%%%%%%%%%%%%%%%%%%%%%%%%%%%%%%%%%%%%%%%%%%%%%%%%%%%%%%%%%%%%%%%%%%%%%%%%%%%%%%%%%%%%%%%%%%%%%%%%%%%%%%%%%%%%%%%%%%%%%%%%%
%%%%%%%%%%%%%%%%%%%%%%%%%%%%%%%%%%%%%%%%%%%%%%%%%%%%%%%%%%%%%%%%%%%%%%%%%%%%%%%%%%%%%%%%%%%%%%%%%%%%%%%%%%%%%%%%%%%%%%%%%%%%%%%%%%%%%%%%
Before showing that the selected index completes the inductive step, we recall the following:
\begin{lemma}[Lemma 3.1 in  \cite{BK20}]
\label{lem:ap:diff}
For a subset $S\subseteq[n]$, let $\AA_{*,S}=\frac{1}{|S|}\sum_{i\in S}\AA_{*,i}$. 
For all $S\subseteq[n]$, $|\AA_{*,S}-\PP_{*,S}|\le \sigma\sqrt{n/|S|}$. 
\end{lemma}

We then show that the algorithm preserves the aforementioned invariant by showing that the unique solution $\AA_{\R_i}$ cannot correspond to one of the vertices of the $k$-simplex that have been found in the first $i$ rounds, thus proving that we find a solution $\AA_{\R_i}$ that corresponds to a new vertex of $\MM$. 
We then show $\AA_{\R_i}$ is close to the new vertex of $\MM$, preserving the inductive hypothesis. 

\begin{lemma}[Recovery Guarantees]
\label{bk:ten:three}
Let $\widehat{\MM}=\MM_{*,\ell_1}\circ\ldots\circ\MM_{*,\ell_r}$ be the $r$ points in the latent $k$-simplex $\MM$ closest to the first $r$ points selected by Algorithm \ref{alg:input_sparsity_simplex}, $\R_1,\ldots,\R_{r}$, respectively. 
Suppose
\[\|\R_{i}-\MM_{*,\ell_i}\|_2\le\frac{300k^4}{\alpha}\frac{\sigma}{\sqrt{\delta}}\] 
for each $i\in[r]$. 
Let $\bu\in\mathbb{R}^d$ be a random unit vector in the space of $\YY^T(\II_d - \PP_r)$, where $\PP_r$ is the orthogonal projection to $\R_1,\ldots,\R_{r}$. 
%Let $\bu=g\YY^T(\II_d - \PP_r)\in\mathbb{R}^d$, where $\PP_r$ is the ortogonal projection to $\R_1,\ldots,\R_{r}$. 
Let
\[\ell_{r+1}=\begin{cases}\argmax_{\ell}\bu\cdot\MM_{*,\ell}\qquad\text{if }\bu\cdot\R_{r+1}\ge 0\\
\argmin_{\ell}\bu\cdot\MM_{*,\ell}\qquad\text{if }\bu\cdot\R_{r+1}<0\end{cases}.\] 
Then
\[\|\R_{r+1}-\MM_{*,\ell_{r+1}}\|_2\le\frac{300k^4}{\alpha}\frac{\sigma}{\sqrt{\delta}}.\] 
\end{lemma}

% \newline
% \noindent\textbf{Lemma~\ref{bk:ten:three}}\textit{ (Recovery Guarantees, Restated).
% Let $\widehat{\MM}=\MM_{*,\ell_1}\circ\ldots\circ\MM_{*,\ell_r}$ be the $r$ points in the latent $k$-simplex $\MM$ closest to the first $r$ points selected by Algorithm \ref{alg:input_sparsity_simplex}, $\R_1,\ldots,\R_{r}$, respectively. 
% Suppose
% \[\|\R_{i}-\MM_{*,\ell_i}\|_2\le\frac{300k^4}{\alpha}\frac{\sigma}{\sqrt{\delta}}\] 
% for each $i\in[r]$. 
% Let $\bu\in\mathbb{R}^d$ be a random unit vector in the space of $\YY^T(\II_d - \PP_r)$, where $\PP_r$ is the orthogonal projection to $\R_1,\ldots,\R_{r}$. 
% %Let $\bu=g\YY^T(\II_d - \PP_r)\in\mathbb{R}^d$, where $\PP_r$ is the ortogonal projection to $\R_1,\ldots,\R_{r}$. 
% Let
% \[\ell_{r+1}=\begin{cases}\argmax_{\ell}\bu\cdot\MM_{*,\ell}\qquad\text{if }\bu\cdot\R_{r+1}\ge 0\\
% \argmin_{\ell}\bu\cdot\MM_{*,\ell}\qquad\text{if }\bu\cdot\R_{r+1}<0\end{cases}.\] 
% Then
% \[\|\R_{r+1}-\MM_{*,\ell_{r+1}}\|_2\le\frac{300k^4}{\alpha}\frac{\sigma}{\sqrt{\delta}}.\] 
% }
\begin{proof}
We consider the case $\bu\cdot\R_{r+1}\ge 0$ as the analysis for the case $\bu\cdot\R_{r+1}<0$ is symmetric. 
Let $\ell_{r+1}=\argmax_{\ell}\bu\cdot\MM_{*,\ell}$. 
By Lemma~\ref{bk:ten:extra}, we have $\ell_{r+1}\notin\{\ell_1,\ldots,\ell_r\}$. 
Thus applying Lemma~\ref{bk:ten:two},
\begin{equation}
\label{bu:lower:decimal}
\bu\cdot\MM_{*,\ell_{r+1}}\ge\frac{0.0989}{k^4}\alpha\max_{\ell}\|\MM_{*,\ell}\|.
\end{equation}
By the Proximate Latent Points assumption, there exists a set $\sigma_{\ell_{r+1}}$ of size $\delta n$ so that $\|\PP_{*,j}-\MM_{*,\ell_{r+1}}\|_2\le\frac{4\sigma}{\sqrt{\delta}}$ for all $j\in \sigma_{\ell_{r+1}}$ so that $\|\PP_{*,\sigma_{\ell_{r+1}}}-\MM_{*,\ell_{r+1}}\|_2\le\frac{4\sigma}{\sqrt{\delta}}$. 
Then by Lemma~\ref{lem:ap:diff}, 
\[\bu\cdot\AA_{*,\sigma_{\ell_{r+1}}}\ge\bu\cdot\PP_{*,\sigma_{\ell_{r+1}}}-\frac{\sigma}{\sqrt{\delta}}\ge\bu\cdot\MM_{*,\ell_{r+1}}-\frac{5\sigma}{\sqrt{\delta}}.\]
By the same reasoning as \ref{bu:astyst}, we have $\|\R_{r+1}-\AA_{*,\sigma_{\ell_{r+1}}}\|_2\le\frac{3\sigma}{\sqrt{\delta}}$ and thus,
\begin{equation}
\label{final:lower}
\bu\cdot\R_{r+1}\ge\bu\cdot\MM_{*,\ell_{r+1}}-\frac{8\sigma}{\sqrt{\delta}}.
\end{equation}
Now for any $a\notin\{\ell_1,\ldots,\ell_{r+1}\}$, Lemma~\ref{bk:ten:two} says
\begin{equation}
\label{uml:upper:one}
\bu\cdot\MM_{*,a}\le\bu\cdot\MM_{*,\ell_{r+1}}-\frac{0.097}{k^4}\alpha\max_{\ell}\|\MM_{*,\ell}\|_2.
\end{equation}
Similarly, for $a\in\{\ell_1,\ldots,\ell_r\}$, we have $\|\mathcal{R}_a-\MM_{*,a}\|\le\frac{300k^4}{\alpha}\frac{\sigma}{\sqrt{\delta}}$ by the inductive hypothesis. 
Since $\bu\cdot\mathcal{R}_a=0$, then
\begin{align*}
\bu\cdot\MM_{*,a}&\le\bu\cdot\mathcal{R}_a+\frac{300k^4}{\alpha}\frac{\sigma}{\sqrt{\delta}}=\frac{300k^4}{\alpha}\frac{\sigma}{\sqrt{\delta}}\\
&\le\bu\cdot\MM_{*,\ell_{r+1}}-\frac{0.0989}{k^4}\alpha\max_{\ell}\|\MM_{*,\ell}\|\\
&+\frac{300k^4}{\alpha}\frac{\sigma}{\sqrt{\delta}}
\end{align*}
by (\ref{bu:lower:decimal}). 
Thus by the Spectrally Bounded Perturbation assumption,
\begin{equation}
\label{uml:upper:two}
\bu\cdot\MM_{*,a}\le\bu\cdot\MM_{*,\ell_{r+1}}-\frac{0.097}{k^4}\alpha\max_{\ell}\|\MM_{*,\ell}\|
\end{equation}
Since $\PP_{*,\R_{r+1}}$ is a convex combination of the columns of $\MM$, there exists a vector $\bw$ such that $\PP_{*,\R_{r+1}}=\MM\bw$.
Then by the same reasoning as \ref{bu:astyst} and Lemma~\ref{lem:ap:diff},
\begin{align*}
\bu\cdot\R_{r+1}&\le\bu\cdot\AA_{*,\R_{r+1}}+\frac{3\sigma}{\sqrt{\delta}}\le\bu\cdot\PP_{*,\R_{r+1}}+\frac{3\sigma}{\sqrt{\delta}}+\frac{4\sigma}{\sqrt{\delta}}\\
&\le w_{\ell_{r+1}}(\bu\cdot\MM_{*,\ell_{r+1}})+\\
& \sum_{a\neq\ell_{r+1}} w_a\left((\bu\cdot\MM_{*,\ell_{r+1}}-\frac{0.097}{k^4}\alpha\max_{\ell}\|\MM_{*,\ell}\|_2\right)\\
&+\frac{4\sigma}{\sqrt{\delta}},
\end{align*}
where the last line follows from decomposing $\MM$ and applying (\ref{uml:upper:one}) and (\ref{uml:upper:two}) to $\MM_{*,a}$ for $a\neq\ell_{r+1}$. 
Hence,
\begin{align*}
\bu\cdot\R_{r+1}&\le\bu\cdot\MM_{*,\ell_{r+1}}-\frac{0.097\alpha\max_{\ell}\|\MM_{*,\ell}\|_2(1-w_{\ell_{r+1}})}{k^4}\\
&+\frac{4\sigma}{\sqrt{\delta}}.
\end{align*}
Combining with (\ref{final:lower}), we have
\[(1-w_{\ell_{r+1}})\max_{\ell}\|\MM_{*,\ell}\|_2\le\frac{12\sigma}{\sqrt{\delta}}\frac{k^4}{0.097\alpha}\le\frac{124k^4}{\alpha}\frac{\sigma}{\sqrt{\delta}}.\]
Thus,
\begin{align*}
\|\PP_{*,\R_{r+1}}-\MM_{*,\ell_{r+1}}\|_2&=\|(w_{\ell_{r+1}}-1)\MM_{*,\ell_{r+1}}\\
& +\sum_{a\neq\ell_{r+1}}w_a\MM_{*,a}\|\\
&\le\sum_{a\neq\ell_{r+1}}w_a\|\MM_{*,\ell_{r+1}}-\MM_{*,a}\|_2\\
&\le2(1-w_{\ell_{r+1}})\max_{\ell}\|\MM_{*,\ell}\|_2\\
&\le\frac{248k^4}{\alpha}\frac{\sigma}{\sqrt{\delta}}.
\end{align*}
Finally from the triangle inequality and Lemma~\ref{lem:ap:diff}, we have
\begin{equation*}
    \begin{split}
        \|\R_{r+1}-\MM_{*,\ell_{r+1}}\|_2& \le\|\R_{r+1}-\PP_{*,\R_{r+1}}\|_2 \\
        &+\|\PP_{*,\R_{r+1}}-\MM_{*,\ell_{r+1}}\|_2\\
&\le\frac{3\sigma}{\sqrt{\delta}}+\frac{248k^4}{\alpha}\frac{\sigma}{\sqrt{\delta}}\\
&\le\frac{300k^4}{\alpha}\frac{\sigma}{\sqrt{\delta}}.
    \end{split}
\end{equation*}
\end{proof}
%%%%%%%%%%%%%%%%%%%%%%%%%%%%%%%%%%%%%%%%%%%%%%%%%%%%%%%%%%%%%%%%%%%%%%%%%%%%%%%%%%%%%%%%%%%%%%%%%%%%%%%%%%%%%%%%%%%%%%%%%%%%%%%%%%%%%%%%
%%%%%%%%%%%%%%%%%%%%%%%%%%%%%%%%%%%%%%%%%%%%%%%%%%%%%%%%%%%%%%%%%%%%%%%%%%%%%%%%%%%%%%%%%%%%%%%%%%%%%%%%%%%%%%%%%%%%%%%%%%%%%%%%%%%%%%%%
%%%%%%%%%%%%%%%%%%%%%%%%%%%%%%%%%%%%%%%%%%%%%%%%%%%%%%%%%%%%%%%%%%%%%%%%%%%%%%%%%%%%%%%%%%%%%%%%%%%%%%%%%%%%%%%%%%%%%%%%%%%%%%%%%%%%%%%%
%%%%%%%%%%%%%%%%%%%%%%%%%%%%%%%%%%%%%%%%%%%%%%%%%%%%%%%%%%%%%%%%%%%%%%%%%%%%%%%%%%%%%%%%%%%%%%%%%%%%%%%%%%%%%%%%%%%%%%%%%%%%%%%%%%%%%%%%

% Lemma[Justification for Significant Singular Values] Consider the Latent Simplex with the Well-Separatedness, Proximate Latent Points and Spectrally Bounded Perturbation assumptions as considered in [BK'20]. Then, any algorithm running in time $o(nnz(A)k)$ that recovers each latent point up to $poly(k/\epsilon))$ also computes a $1+\epsilon$ spectral low rank approximation for $A$ in the same running time.

\section{Connection to Spectral Low-Rank Approximation}
\label{app:mixed:reduction}
In this section, we show that learning a latent simplex is closely related to computing a spectral low-rank approximation. Spectral low-rank approximation is a fundamental primitive for algorithm design and numerical linear algebra and the best known algorithm for computing a $(1+\epsilon)$-approximation is $O(\nnz(\AA)\cdot k)$  \cite{musco2015randomized}. A major open question in randomized linear algebra is to determine whether the dependence on $k$ in the running time is necessary for spectral low-rank approximation. 

We show that for a candidate hard distribution over the input, determined by a Stochastic Block Model (with appropriate parameters) satisfying Well-Separateness\ref{a1}, Proximate Latent Points\ref{a2} and Spectrally Bounded Perturbations\ref{a3}, an algorithm for learning a latent simplex requiring $o(\nnz(\AA)\cdot k)$ time also recovers a spectral low-rank approximation for the input. One way to interpret this statement is that improving the running time for learning a latent simplex under the same assumptions as  \cite{BK20} would likely lead to a major algorithmic breakthrough for spectral low-rank approximation.

%\begin{theorem}[Spectral LRA to Latent Simplex]
%Given $k \in [n]$, let $\S_1 , \S_2 \ldots, \S_k$ be a partition of $[n]$ such that for all $\ell \in [k]$, $|\S_\ell| = n/k$. Consider a stochastic block model with $k$ communities, $\S_1, \ldots, \S_k$ such that for all $i \in \S_{\ell}$ and $j \in \S_{\ell'}$, the probability of an edge $(i,j)$ is $p= \poly(k)/n^{7/8}$ when $\ell = \ell'$ and $q = p/10$ otherwise. Let $\AA$ be a matrix drawn from the aforementioned model such that $\AA_{i,j} =1$ if there exists an edge between $(i,j)$ and $0$ otherwise. Given $\epsilon >0$, any algorithm that learns the simplex also recovers a rank $k$ matrix $\BB$ such that $\|\AA - \BB \|^2_2 \leq (1+\epsilon)\|\AA - \AA_k \|^2_2$. 
%\end{theorem}

\begin{theorem}[Spectral LRA to Latent Simplex]
Given $k \in [n]$, let $\S_1 , \S_2 \ldots, \S_k$ be a partition of $[n]$ such that for all $\ell \in [k]$, $|\S_\ell| = n/k$. Consider a stochastic block model with $k$ communities, $\S_1, \ldots, \S_k$ such that for all $i \in \S_{\ell}$ and $j \in \S_{\ell'}$, the probability of an edge $(i,j)$ is $p=\poly(k)/n^{1/8}$ when $\ell = \ell'$ and $q = p/10$ otherwise. Let $\AA$ be a matrix drawn from the aforementioned model such that $\AA_{i,j} =1$ if there exists an edge between $(i,j)$ and $0$ otherwise. Then any algorithm that learns the simplex also recovers a rank $k$ matrix $\BB$ such that $\|\AA - \BB \|^2_2 \leq \|\AA - \AA_k \|^2_2+\frac{1}{n^{1/3}}\|\AA-\AA_k\|^2_F$. 
\end{theorem}
\begin{proof}
Let $\PP_{\BB}$ be the projection matrix onto the column span of the output matrix $\BB$.  
We show that $\AA-\PP_{\BB}$ is a good mixed spectral-Frobenius low-rank approximation to $\AA$. 
\begin{align*}
\|\AA-\PP_{\BB}\AA\|_2&\le\|\AA-\PP+\PP\|_2\|\II-\PP_{\BB}\|_2\\
&\le\|\AA-\PP\|_2\|\II-\PP_{\BB}\|_2+\PP\|_2\|\II-\PP_{\BB}\|_2\\
&\le\|\AA-\PP\|_2+\|\PP\|_2\|\II-\PP_{\BB}\|_2.
\end{align*}
From the definition of $\sigma$, we have $\|\AA-\PP\|_2\le\sigma\sqrt{n}$. 
For the specific stochastic block model, we have $\sigma\le\sqrt{p(1-p)}$, e.g., see~ \cite{pranjalnotes}. 
Moreover, the algorithm of  \cite{BK20} guarantees specifically in their Theorem 7.2 that $\|\II-\PP_{\BB}\|_2\le\frac{C_1 k^{4.5}d^{1/8}}{n^{1/4}}$ for some constant $C_1>0$. 
Since $\|\PP\|_F\ge\|\PP\|_2$ and $\|\PP\|_F^2\le C_2 p^2 nd$ for some constant $C_2>0$ with high probability, then we have
\begin{align*}
\|\AA-\PP_{\BB}\AA\|_2&\le\sqrt{p(1-p)n}+\frac{C_1 k^{4.5}d^{1/8}\sqrt{C_2 p^2 nd}}{n^{1/4}}\\
&\le\sqrt{pn}+C_1 pk^{4.5}d^{5/8}\sqrt{C_2}n^{1/4}.
\end{align*}
On the other hand, we have $\|\AA-\AA_k\|^2_F\ge\|\AA\|^2_F-k\|\AA\|^2_2$. 
As before, we have $\|\PP\|_2\le p\sqrt{C_2 nd}$, so that  
\[\|\AA\|_2\le\|\PP\|_2+\|\AA-\PP\|_2\le p\sqrt{C_2 nd}+\sqrt{p(1-p)n}.\] 
Moreover, we have $\|\AA\|_F\ge C_3\sqrt{qnd}$ for some constant $C_3>0$ with high probability. 
Hence for $q>C_4 p^2$ with a sufficiently high constant $C_4$, we have 
\[\|\AA-\AA_k\|_F^2\ge C_5 qnd,\]
for some $C_5>0$.  
Let $p=O(q)$ and $d=n^{1/C}$ for some constant $C\ge 3$ so that $k^{4.5}d^{5/8}=o(n^{1/4})$.  
Since $\|\AA-\PP_{\BB}\AA\|^2_2\le C_6pn$ for some constant $C_6$, then
\begin{align*}
\|\AA-\PP_{\BB}\AA\|^2_2&\le C_6pn\le \frac{C_5}{n^{1/C}}qnd=O\left(\frac{1}{n^{1/C}}\right)\|\AA-\AA_k\|_F^2\\
&\le\|\AA-\AA_k\|_2^2+O\left(\frac{1}{n^{1/C}}\right)\|\AA-\AA_k\|_F^2.
\end{align*}
Taking $C=3$ gives the desired claim. 
\end{proof}
\section{Empirical Evaluation}
In this section, we describe a series of experiments that demonstrate the advantage of our algorithm, performed in Python 3.6.9 on an Intel Core i7-8700K 3.70 GHz CPU with 12 cores and 64GB DDR4 memory, using an Nvidia Geforce GTX 1080 Ti 11GB GPU, on both synthetic and real-world data. 
Whereas previous work requires computing the top $k$ subspace as a pre-processing step, our main improvement is that we only require a crude approximation. 
Thus we compared the running times for finding the top $k$ subspace as required by  \cite{BK20} to finding a mixed spectral-Frobenius approximation using an input sparsity algorithm, as required by our algorithm. 
For the former, we use the \texttt{svds} method from the sparse scipy linalg package optimized by LAPACK. 
For the latter,  \cite{cohen2015dimensionality, cohenmm17} show that using a sparse CountSketch matrix~ \cite{clarkson2013low,meng2013low,NelsonN13}, i.e., a matrix with $O(k^2)$ columns and a single nonzero entry in each row that is in a random location and is a random sign, suffices to obtain a mixed spectral-Frobenius guarantee; we evaluate such a matrix with exactly $k^2$ columns. 
Across all parameters and datasets, the input sparsity procedure used by our algorithm significantly outperforms the optimized power iteration methods required by~ \cite{BK20}.

\textbf{Synthetic Data.}
Since our theoretical results are most interesting when $k\ll d\ll n$, we set $n=50000$, $d=1000$, $k\in\{20,50,100\}$ and generate a random $d\times n$ matrix $\AA$ that consists of independent entries that are each $1$ with probability $p\in\left\{\frac{1}{500},\frac{1}{2000},\frac{1}{5000}\right\}$ and $0$ with probability $1-p$. 
In Figure~\ref{fig:synthetic}, we report the average running time of both algorithms, among $5$ independent runs for each choice of $p$ and $k$. 
%The input sparsity procedure used by our algorithm significantly out-performs the optimized power iteration methods required by~ \cite{BK20} across all parameters. 

\begin{figure}[!htb]
\centering
\begin{tabular}{|c|c|c|c|}\hline
Mean Runtime of Algorithms across Parameters
 & $p=1/500$ & $p=1/2000$ & $p=1/5000$  \\\hline
Top $k$ Subspace, $k=20$ & 35.056s & 29.725s & 16.45s \\\hline
Input Sparsity Approximation, $k=20$ & 0.595s & 0.329s & 0.83s  \\\hline
Top $k$ Subspace, $k=50$ & 56.146s & 54.613s & 53.213s \\\hline
Input Sparsity Approximation, $k=50$  & 0.658s & 0.657s & 0.434s \\\hline
Top $k$ Subspace, $k=100$ & 78.420s & 79.410s & 71.424s  \\\hline
Input Sparsity Approximation, $k=100$  & 0.501s & 0.387s & 0.440s \\\hline
\end{tabular}
\caption{Mean runtime comparison of algorithms across parameters on synthetic data.}\label{fig:synthetic}
\end{figure}

\textbf{Social Networks.}
We also evaluate the algorithms on the \texttt{email-Eu-core network} dataset of interactions across email data between individuals from a large European research institution~ \cite{YinBLG17,LeskovecKF07} and the \texttt{com-Youtube} dataset of friendships on the Youtube social network~ \cite{YangL15}, both accessed through the Stanford Network Analysis Project (SNAP). 
In the former, there are $n=d=1005$ nodes in the adjacency matrix over $25571$ total edges, forming $k=42$ communities. 
In the latter, there are $1134890$ nodes with $8385$ communities, from which we extract a $d\times n$ matrix with $n=100000$, $d=1000$ to represent a bipartite graph, as described in both Section~\ref{section:MMBM} and  \cite{BK20}. 
In Figure~\ref{fig:snap}, we report the running time of both algorithms across each dataset among choices of $k\in\{20,50,100\}$. 
We observe that the resulting matrix has sparsity roughly $1000$, which is consistent with $p\approx\frac{1}{n}$ and is much less than the sparsity parameters tested in our synthetic data. 
%Again the input sparsity subroutine used by our algorithm is significantly faster than the optimized power iteration methods required by~ \cite{BK20} across all parameters. 

\begin{figure}[!htb]
\centering
\begin{tabular}{|c|c|c|}\hline
%\multicolumn{3}{|c|}{Mean Runtime of Algorithms across Parameters}\\\hline
 & \texttt{email-Eu-core network} & \texttt{com-Youtube} \\\hline
Top $k$ Subspace, $k=20$ & 0.387s & 5.713s  \\\hline
Input Sparsity Approximation, $k=20$  & 0.005s & 0.379s \\\hline
Top $k$ Subspace, $k=50$ & 0.556s & 16.711s  \\\hline
Input Sparsity Approximation, $k=50$  & 0.003s & 0.373s  \\\hline
Top $k$ Subspace, $k=100$ & 1.281s & 41.788s  \\\hline
Input Sparsity Approximation, $k=100$  & 0.003s & 0.366s  \\\hline
\end{tabular}
\caption{Mean runtime comparison of algorithms across parameters on real-world data.}\label{fig:snap}
\end{figure}

\begin{figure*}[!hbt]
\centering
\begin{subfigure}[b]{0.3\textwidth}
\begin{tikzpicture}[scale=0.4]
\begin{axis}[
    ylabel={Least Squared Loss},
    xlabel={Number of communities ($k$), with $\delta n=10$},
    ymin=0, ymax=2100,
    xmin=0, xmax=30,
    xtick={0,5,10,15,20,25,30},
	ytick={0,300,600,900,1200,1500,1800,2100},
		ymajorgrids=true,
    grid style=dashed,
]
\addplot[
    color=purple,
    mark=*,
    ]
    coordinates {(1, 2014.7254782872687)
(2, 2025.9119362640527)
(3, 1911.2134822497026)
(4, 1893.4310531615131)
(5, 1922.9062269456049)
(6, 1815.0637519815646)
(7, 1714.5213413913075)
(8, 1867.8550831636837)
(9, 1901.863186292768)
(10, 1700.4486620928146)
(11, 1618.0201773994067)
(12, 1736.763378831846)
(13, 1680.0492183708029)
(14, 1631.466497174453)
(15, 1751.6324057507757)
(16, 1477.6052788714635)
(17, 1514.3682936498972)
(18, 1450.565007283833)
(19, 1457.0594119244199)
(20, 1474.9004706252442)
(21, 1436.7474910005071)
(22, 1450.9128720062201)
(23, 1456.5103437735249)
(24, 1499.1419759602691)
(25, 1404.4004307536693)
(26, 1294.3271569927911)
(27, 1288.4845221126161)
(28, 1340.1522897223153)
(29, 1324.983145410981)
(30, 1235.2150388069658)
};
\addplot[
    color=blue,
    mark=triangle*,
    ]
    coordinates {(1, 2086.1216216215885)
(2, 2096.8611111111454)
(3, 2042.7238507000893)
(4, 2005.6079711520208)
(5, 2047.2225300652544)
(6, 1883.316619790038)
(7, 1995.3816195847799)
(8, 2023.5440355203998)
(9, 2064.6082180845374)
(10, 2012.1072625007523)
(11, 1853.3331629954268)
(12, 1977.036401118668)
(13, 1881.1612870089348)
(14, 1890.3832576586151)
(15, 1981.2902839835335)
(16, 1899.0938694174436)
(17, 1731.4820203719378)
(18, 1770.692055104503)
(19, 1874.7519214361403)
(20, 1816.8227432140279)
(21, 1846.0096891633532)
(22, 1893.8181736616675)
(23, 1673.8907524907193)
(24, 1940.3368919067618)
(25, 1888.3985603171138)
(26, 1707.3621583929018)
(27, 1691.2137266750224)
(28, 1670.0944733349843)
(29, 1728.7128130591716)
(30, 1589.520939550172)};
\end{axis}
\end{tikzpicture}
%\caption{Step size $\eta=10^{-4}$, Regularization $1$}
%\label{fig:svm:madelone:obj}
\end{subfigure}
%%%%%%%%%%%%%%%%%%%%%%%%%%%%%%%%%%%%%%%%%%%%%%%%%%%%%%%%
\begin{subfigure}[b]{0.3\textwidth}
\begin{tikzpicture}[scale=0.4]
\begin{axis}[
    ylabel={Least Squared Loss},
    xlabel={Smoothening parameter $(\delta n)$, with $k=20$},
    ymin=0, ymax=2100,
    xmin=0, xmax=30,
    xtick={0,5,10,15,20,25,30},
	ytick={0,300,600,900,1200,1500,1800,2100},
		ymajorgrids=true,
    grid style=dashed,
]
\addplot[
    color=purple,
    mark=*,
    ]
    coordinates {(1, 1557.403180338126)
(2, 1497.1178640105773)
(3, 1642.677263908932)
(4, 1518.349556960163)
(5, 1428.1843927805858)
(6, 1581.1345861808525)
(7, 1434.8988540305552)
(8, 1484.0075656154604)
(9, 1419.5289806976652)
(10, 1446.7897350250526)
(11, 1499.2488088475009)
(12, 1461.4308814941567)
(13, 1446.3896123275695)
(14, 1478.044782172429)
(15, 1388.7626992348339)
(16, 1454.7938989028248)
(17, 1424.4773890005783)
(18, 1393.8160337932254)
(19, 1312.2529991412496)
(20, 1364.8122482984952)
(21, 1428.4236671362214)
(22, 1394.5671177449215)
(23, 1417.8273239683963)
(24, 1359.4483362252231)
(25, 1341.0189531471065)
(26, 1425.2230597546402)
(27, 1403.3644878345021)
(28, 1461.1825248588079)
(29, 1473.6385073252566)
(30, 1395.6454698215944)
};
\addplot[
    color=blue,
    mark=triangle*,
    ]
    coordinates {(1, 2024.1309523809525)
(2, 1785.5186420028178)
(3, 2063.3172484050951)
(4, 1869.5345050858077)
(5, 1820.9757433650518)
(6, 1986.6889210971497)
(7, 1818.0939787929497)
(8, 1800.8771393755458)
(9, 1782.1409097476437)
(10, 1903.5119536837415)
(11, 1813.1207459207094)
(12, 1767.1007709537489)
(13, 1855.3105901172328)
(14, 1883.9342612417045)
(15, 1639.4275804572821)
(16, 1791.1703560984631)
(17, 1837.1518540680795)
(18, 1682.5545220391434)
(19, 1784.5743215955035)
(20, 1695.8333646101896)
(21, 1670.4766038831065)
(22, 1728.4531791088302)
(23, 1727.9611589308322)
(24, 1738.3461188457047)
(25, 1583.6883769277013)
(26, 1722.6624640475868)
(27, 1652.670958720518)
(28, 1675.5784995592053)
(29, 1772.8772230115485)
(30, 1693.3521849050028)};
\end{axis}
\end{tikzpicture}
%\caption{}
%\label{fig:svm:madelone:val}
\end{subfigure}
%%%%%%%%%%%%%%%%%%%%%%%%%%%%%%%%%%%%%%%%%%%%%%%%%%%%%%
\begin{subfigure}[b]{0.3\textwidth}
\begin{tikzpicture}[scale=0.4]
\begin{axis}[
    ylabel={Runtime},
    xlabel={Number of communities ($k$), with $\delta n=10$},
    ymin=0, ymax=9,
    xmin=0, xmax=30,
    xtick={0,5,10,15,20,25,30},
	ytick={0,1,2,3,4,5,6,7,8,9},
		ymajorgrids=true,
    grid style=dashed,
]
\addplot[
    color=purple,
    mark=*,
    ]
    coordinates {(1, 0.22750234603881836)
(2, 0.1558377742767334)
(3, 0.30481934547424316)
(4, 0.20823931694030762)
(5, 0.40145087242126465)
(6, 0.490389347076416)
(7, 0.399599552154541)
(8, 0.762070894241333)
(9, 0.6448793411254883)
(10, 0.6823718547821045)
(11, 0.7465627193450928)
(12, 0.8870975971221924)
(13, 0.8516838550567627)
(14, 1.0694372653961182)
(15, 0.9315621852874756)
(16, 1.6987242698669434)
(17, 1.5678651332855225)
(18, 1.5545930862426758)
(19, 1.5163671970367432)
(20, 1.7052829265594482)
(21, 1.5250561237335205)
(22, 2.48884916305542)
(23, 1.6044921875)
(24, 2.0171010494232178)
(25, 2.793992757797241)
(26, 2.059593677520752)
(27, 3.4624221324920654)
(28, 1.8323147296905518)
(29, 2.946528911590576)
(30, 1.808840274810791)
};
\addplot[
    color=blue,
    mark=triangle*,
    ]
    coordinates {(1, 0.9919402599334717)
(2, 1.3288207054138184)
(3, 2.18870210647583)
(4, 2.5847604274749756)
(5, 2.3886911869049072)
(6, 2.2748959064483643)
(7, 2.4390759468078613)
(8, 3.208632230758667)
(9, 3.2358438968658447)
(10, 3.2942023277282715)
(11, 3.6183221340179443)
(12, 3.240051031112671)
(13, 4.043457269668579)
(14, 5.037900924682617)
(15, 3.9972455501556396)
(16, 4.6700756549835205)
(17, 3.726712465286255)
(18, 4.720227003097534)
(19, 5.218005657196045)
(20, 4.724025249481201)
(21, 5.837247610092163)
(22, 5.865950107574463)
(23, 5.870581865310669)
(24, 6.671548843383789)
(25, 7.512039422988892)
(26, 6.22937536239624)
(27, 8.420896053314209)
(28, 6.7569708824157715)
(29, 7.518426418304443)
(30, 6.6563074588775635)};
\end{axis}
\end{tikzpicture}
%\caption{Step size $\eta=10^{-4}$, Regularization $1$}
%\label{fig:svm:madelone:obj}
\end{subfigure}
\caption{Comparison of least squares loss by power iteration algorithm (in blue triangles) and by our algorithm (in red circles), over various ranges of the parameter $k$ with smoothening parameter $\delta n=10$, and over various ranges of $\delta n$ with $k=20$, on the \texttt{com-Youtube} dataset. Also runtime comparison over a range of $k$, with $\delta n=10$.}
\label{fig:experiments}
\end{figure*}
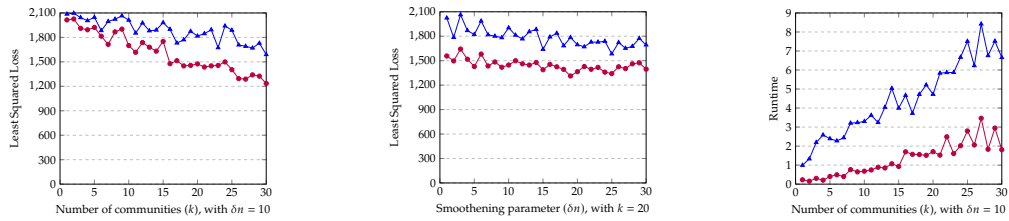

Finally, we consider a full end-to-end implementation comparing the runtime and least squares loss of the top $k$ subspace algorithm and our input sparsity approximation algorithm over various ranges of the parameter $k$ and smoothening parameter $\delta n$ on the \texttt{com-Youtube} dataset, from which we randomly extract an $n\times d$ matrix, with $n=20000$ and $d=1000$ to represent a bipartite graph.
Our results in Figure~\ref{fig:experiments} show that our algorithm not only significantly outperforms the top $k$ subspace algorithm in runtime, but also produces solutions with lower least squared loss.

\subsubsection*{Acknowledgments}
A.B., and D. W. were supported by the Office of Naval Research (ONR) grant N00014-18-1-2562, and the
National Science Foundation (NSF) Grant No. CCF-1815840. D.W and S.Z were supported by 
National Institute of Health (NIH) grant 5R01 HG 10798-2 and a Simons Investigator Award. 

\bibliography{polytope,jmlr_bib,nmf}
\bibliographystyle{alpha}

\end{document}